\documentclass{article} 
\usepackage{iclr2027_conference,times}

\usepackage{amsmath,amsfonts,bm}

\def\eqref#1{equation~\ref{#1}}

\def\1{\bm{1}}

\DeclareMathAlphabet{\mathsfit}{\encodingdefault}{\sfdefault}{m}{sl}
\SetMathAlphabet{\mathsfit}{bold}{\encodingdefault}{\sfdefault}{bx}{n}

\usepackage{hyperref}
\usepackage{url}
\usepackage{graphicx}
\usepackage{booktabs}
\usepackage{flafter}
\usepackage{ulem}

\title{Hierarchical Spatiotemporal World Modeling with Content-Adaptive Spatial Addressing}
\author{
Tiantian Li$^{1}$ \quad
Junfeng Lu$^{2}$ \quad
Yifei Wang$^{1}$ \quad
Zhaoyang Zhang$^{1}$ \quad
Lei Wang$^{3,*}$ \quad
Tianyi Qian$^{1,*}$ \\[0.5em]
{\normalfont $^{1}$QiYuanLab} \quad
{\normalfont $^{2}$Peking University} \quad
{\normalfont $^{3}$Academy of Military Sciences} \\[0.3em]
{\normalfont $^{*}$Corresponding authors}
}

\iclrfinalcopy

\begin{document}

\maketitle
\fancyhf{}
\fancyfoot[C]{\thepage}
\renewcommand{\headrulewidth}{0pt}
\renewcommand{\footrulewidth}{0pt}
\pagestyle{fancy}
\thispagestyle{fancy}

\begin{abstract}
 Long-horizon world modeling in partially observable environments requires compact memories that preserve spatial structure despite accumulated localization drift. Inspired by 
 hippocampal episodic and spatial memory and the multi-periodic coding of entorhinal grid cells, we propose HiSTFormer, a hierarchical spatiotemporal world model, which adopts a layer-wise organization of local temporal RoPE and multi-periodic spatial RoPE, together with content-conditioned address offsets, to capture recent temporal context and long-range spatial structure. A fixed-capacity memory keeps KV storage and per-step attention cost bounded as the sequence grows. On MemoryMaze and HM3D, HiSTFormer outperforms strong world-model baselines within the training sequence length and generalizes substantially better to prediction horizons far beyond those seen during training, while using a compact model. The hierarchical spatiotemporal design remains effective across deeper Transformer stacks, with consistent gains as depth increases. These results show that 
 hierarchical spatiotemporal organization and content-adaptive spatial addressing provide effective inductive biases
  for efficient long-horizon world modeling.  
\end{abstract}

\section{Introduction}
First-person world models must predict local dynamics from consecutive observations and actions while maintaining scene consistency over long interactions using spatial information from past trajectories. Studies such as MemoryMaze and TECO highlight the challenges of long-term memory and visual consistency during scene revisits~\citep{pasukonis2022memorymaze,yan2023TECO}. \textbf{How to organize and effectively utilize spatiotemporal information has therefore become an important question for long-horizon visual prediction.}

Recent studies have increasingly recognized the importance of spatial memory and explored diverse approaches. \textbf{Geometry-based methods} maintain historical scene information through explicit or latent 3D representations. SpMem, 3D Persistent World Model, PERSIST, and Spatia employ 3D maps, geometric indexing, or persistent scene states to maintain spatial consistency~\citep{wu2025spatialmemory,zhou2025persistent,garcin2026persist,zhao2026spatia}. LSM-World and MosaicMem further associate visual latent features with 3D space, recovering historical content through viewpoint transformations and feature alignment~\citep{wang2026latentspatialmemory,yu2026mosaicmem}. While these approaches provide cross-view spatial constraints, they typically require additional geometric estimation, 3D mapping, or spatial memory maintenance.

Another line of research focuses on \textbf{establishing spatial associations through pose information and latent memory mechanisms}. ReMind introduces camera-aware RoPE to establish spatial relationships across time~\citep{xu2026teaching}; ReWorld combines pose-aware attention with a pose-indexed Landmark Bank for long-term memory retrieval~\citep{chen2026reworld}; WorldCrafter retrieves relevant content from implicit 3D-aware memory conditioned on target camera poses~\citep{yu2026worldcrafter}; and MemoryVLN integrates pose, visual, and geometric features to enhance historical memory retrieval for navigation~\citep{wei2026memoryvln}. Other studies explore the division of computational roles across memory mechanisms. CoME combines short-term dynamics, long-term experience, and spatial information through specialized memory experts~\citep{stapf2026CoME}, while the concurrent work LOCI integrates geometry-conditioned recurrent memory and historical KV attention across different Transformer blocks~\citep{xia2026loci}. These approaches demonstrate diverse ways to incorporate spatial information into memory retrieval and organization. However, \textbf{how temporal and spatial relationships should be coordinated across Transformer depth remains insufficiently explored.}

Biological neural systems provide insights into the organization and use of spatial information. Entorhinal grid cells exhibit multi-scale spatial representations associated with path integration, while the hippocampus integrates spatial, temporal, and episodic information~\citep{hafting2005GridCell,moser2014gridcells,eichenbaum2017integration}. Computational models such as TEM, Hippoformer, and HPC-MEC have further explored associations between spatial structure and experience-specific content~\citep{whittington2020tem,kymn2024binding,zhang2026structure,li2026hippoformer}.

We investigate two further questions. First, \textbf{local temporal integration and long-range spatial retrieval have distinct information requirements}: the former relies on consecutive observations and actions, while the latter accesses spatially related experiences across time. How to coordinate them within a Transformer remains underexplored. Second, \textbf{spatial addresses do not necessarily correspond one-to-one with visual content}: visual features may encode information from multiple locations, and fixed camera-centered addresses may limit retrieval accuracy, motivating content-adaptive spatial addressing.

Additionally, GridPE, nD-RoPE, M-RoPE, and VideoRoPE provide technical foundations for encoding multidimensional spatiotemporal relationships in Transformers~\citep{li2024gridpe,li2026nd,wang2024qwen2,wei2025videorope}.

Building on these insights, we introduce \textbf{HiSTFormer}, a hierarchical Transformer for long-horizon visual prediction. Using grid-cell-inspired multi-periodic spatial encoding, HiSTFormer focuses on \textbf{hierarchical spatiotemporal organization} and \textbf{content-adaptive spatial addressing}. First, HiSTFormer organizes temporal and spatial processing into successive stages. Temporal layers use standard temporal RoPE to integrate consecutive observations and actions, capturing local dynamics~\citep{su2021roformer}. Spatial layers then employ multi-periodic spatial RoPE to retrieve historical experiences through path-integrated spatial addresses, enabling scene recovery across temporal gaps. Second, HiSTFormer introduces \textbf{content-conditioned per-head spatial address offsets}, allowing individual attention heads to adapt their retrieval addresses to the represented content. A fixed-capacity structured memory further supports long-range retrieval without unbounded cache growth.

On long-horizon visual prediction in MemoryMaze, HiSTFormer, trained on 320-frame sequences, generalizes reliably to 1,000-frame rollouts and outperforms the compared baselines both within and beyond the training range. Further analyses show that hierarchical temporal–spatial encoding outperforms single-type and within-layer joint encodings, with increasing benefits as Transformer depth grows. Moreover, content-conditioned offsets introduce only \textbf{0.05M additional parameters}, yet improve historical memory retrieval and learn spatial adjustments that tend to align with salient walls.

Overall, these results demonstrate that effective spatiotemporal organization and content-adaptive addressing can improve long-horizon prediction and memory utilization in world models.

\section{Method}
\begin{figure*}[t]
    \centering
    \includegraphics[width=\textwidth]{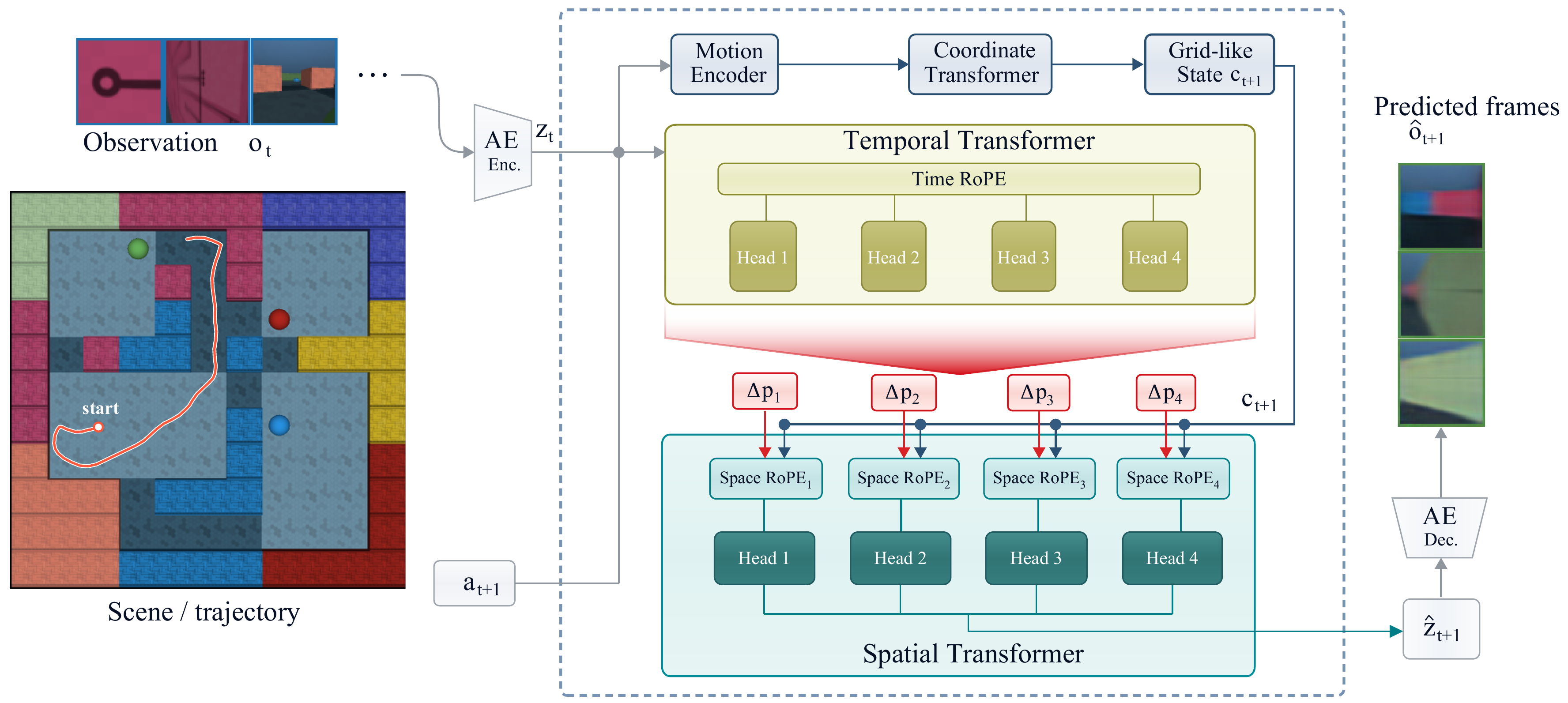}
    \caption{Overview of HiSTFormer. Observation frames are encoded into compact
    visual representations, while the motion encoder and coordinate
    transformation update a path-integrated grid-like spatial state. The
    temporal Transformer first extracts local context using temporal RoPE.
    Content-conditioned per-head offsets then adjust the grid-like spatial
    addresses used by the spatial Transformer to retrieve historical
    experience. The resulting latent representations are decoded
    autoregressively into future frames.}
    \label{fig:model_overview}
\end{figure*}

HiSTFormer is a hierarchical spatiotemporal world model for long-horizon visual prediction, as illustrated in Figure~\ref{fig:model_overview}. At time $t$, the observation $o_t$ is encoded into a visual representation $\mathbf{z}_t$. Given $\mathbf{z}_t$ and the next action $\mathbf{a}_{t+1}$, the temporal Transformer integrates local observations and actions into a context-aware experience representation. The spatial Transformer then retrieves and integrates relevant historical experiences to predict $\hat{\mathbf{z}}_{t+1}$.

Spatial addresses are maintained through path integration. A motion encoder estimates the displacement $\hat{s}_{t+1}$ from $\mathbf{a}_{t+1}$ and the visual representation $\mathbf{z}_{t+1}$ when available. The displacement is transformed into world coordinates using the ground-truth heading and integrated into multi-periodic spatial phases $\mathbf{c}_{t+1}$ for spatial RoPE. Additionally, a lightweight MLP predicts content-conditioned, per-head offsets $\boldsymbol{\Delta p}^{h}$ from the temporal Transformer output, adapting spatial attention addresses to the content represented by each experience.

\subsection{Hierarchical Spatiotemporal Transformer}
Temporal and spatial RoPE are organized hierarchically. The temporal layers
extract local contextual information from recent actions and observations,
whereas the spatial layers retrieve historical information according to
relative spatial relationships. At each step, the model takes the current
visual representation $\mathbf{z}_t$ and the next action $\mathbf{a}_{t+1}$ as
input and produces the predicted $\hat{\mathbf{z}}_{t+1}$. During
observation, $\mathbf{z}_t$ is encoded from the ground-truth observation;
during open-loop prediction, the model instead uses the predicted
representation $\hat{\mathbf{z}}_t$ from the previous step.

Both layer types use standard pre-layer-normalization (Pre-LN) Transformer
blocks composed of multi-head self-attention and feed-forward networks, with a
residual connection around each sublayer. Each attention module uses four
heads, with 256-dimensional query, key, and value representations per head.

The temporal layers combine standard temporal RoPE ~\citep{su2024roformer} with sliding-window
attention and, by default, attend to the 32 most recent tokens. The spatial
layers use a bounded KV cache containing entries from the 32 most recent time steps and up to 96 earlier time steps. As shown in
Appendix~\ref{app:memory_management}, performance is insensitive to the
specific memory-selection rule provided that broad temporal coverage is
maintained.

Each retained spatial-layer entry already contains the local context extracted
by the temporal layers and can therefore carry information beyond that of a single
frame.

\subsection{Spatial Representation RoPE}
The coordinate transformer rotates the predicted displacement from the egocentric to the world coordinate frame using the ground-truth head direction. A causal temporal convolution estimates the motion displacement 
$\hat{s}_{t+1}$ from $a_{t+1}$ alone during imagination, or from both $a_{t+1}$ and $z_{t+1}$ during observation.
Architectural and training details are provided in Appendix~\ref{app:motion_encoder}.

Let the ground-truth head direction be represented by the unit vector
$\mathbf{d}_t=[\cos\theta_t,\sin\theta_t]^{\top}$. Given the predicted
displacement $\hat{s}_t$, the corresponding displacement in world coordinates
is $\Delta\mathbf{r}_t=\hat{s}_t\mathbf{d}_{t-1}$

Position is represented using structured multi-period Fourier phases. The model uses $p$ spatial periods spaced logarithmically in scale and $w$ uniformly distributed plane-wave directions for each period, where $w$ is a multiple of three. The wave vectors are organized into groups of three directions separated by $120^\circ$, following the directional structure of hexagonal spatial codes.
Each plane wave maintains an independent complex phase and corresponding
sine/cosine components, yielding $D_r=2pw$ rotary dimensions.

For each wave vector $\mathbf{k}$, the corresponding unit-complex phase is updated through path integration:
\begin{equation}
c_{t}(\mathbf{k})
=
c_{t-1}(\mathbf{k})
\exp\!\left(
\mathrm{i}\,\mathbf{k}^{\top} \boldsymbol{\Delta r}_t
\right),
\qquad
c_{0}(\mathbf{k})=1.
\end{equation}

Using each unit-complex phase to rotate a two-dimensional channel pair of the
query and key yields the spatial RoPE. This operation explicitly introduces the
relative displacement between estimated positions into the attention
computation, allowing the model to exploit spatial relationships directly.
Combining multiple spatial periods provides position-phase information at
multiple scales, supporting both local positional discrimination and
disambiguation over larger spatial ranges.

The head direction $\theta_t$ is repeated across
16 channel pairs and combined with the position phases above to form the
spatial RoPE.

\subsection{Content-Conditioned Spatial Addressing}
A single observation frame may contain content associated with multiple
physical locations, while a local experience representation aggregates
information across multiple frames. To adapt spatial addressing to the
represented content, we predict a two-dimensional positional offset $\boldsymbol{\Delta p}^h$ for each
spatial-attention head $h$ from the output of the temporal layers.

Specifically, path integration produces the shared base phase
$c_{t,g,j}$ for all spatial-attention heads. The content-conditioned offset
then generates a head-specific spatial address through an additional rotation:
\begin{equation}
\widetilde{c}_{t}^{\,h}(\mathbf{k})
=
c_{t}(\mathbf{k})
\exp\!\left(
\mathrm{i}\,\mathbf{k}^{\top}\boldsymbol{\Delta p}^h
\right).
\end{equation}
This design allows different attention heads to learn distinct
content-conditioned spatial associations for the same experience
representation.

\subsection{Training Objective}
Training is performed on 320-frame sequences divided into chunks of $K=16$ frames. After observing each chunk,
the model autoregressively predicts the next $K$ frames using the observation
memory accumulated up to the current point, together with the actions and head
directions from the subsequent chunk. The model then observes the ground-truth frames of the predicted chunk, and this observe–predict cycle repeats throughout the sequence. The training objective consists of an
image-reconstruction loss $\mathcal{L}_{\mathrm{recon}}$, an open-loop prediction loss $\mathcal{L}_{\mathrm{pred}}$, a pairwise spatial representation loss $\mathcal{L}_{\mathrm{pair}}$, and a teacher--student motion loss $\mathcal{L}_{\mathrm{student}}$. The complete objective is
\begin{equation}
\mathcal{L}
=
\mathcal{L}_{\mathrm{recon}}
+\mathcal{L}_{\mathrm{pred}}
+\mathcal{L}_{\mathrm{student}}
+0.03\mathcal{L}_{\mathrm{pair}}.
\end{equation}

Let $D_{\psi}$ denote visual decoder, the reconstruction loss for an observed frame $t$ is
\begin{equation}
\mathcal{L}_{\mathrm{recon}}^{t}
=
\operatorname{MSE}\!\left(
D_{\psi}\!\left(\mathbf{z}_t\right),
o_t
\right).
\end{equation}

The open-loop prediction loss is defined as the average reconstruction error over all predicted frames, where  $T=320$ is the training sequence length.
\begin{equation}
\mathcal{L}_{\mathrm{pred}}
=
\frac{\sum_{t=K+1}^{T}
\operatorname{MSE}\!\left(
D_{\psi}\!\left(\hat{\mathbf{z}}_{t}\right),
o_t
\right)}{T-K},
\end{equation}

Without ground-truth position supervision, we use visual similarity to encourage spatial consistency in path-integrated representations. The pair loss is applied only to observed trajectories, supervising the visual-action motion encoder. We select frame pairs separated by more than $K$ frames with a circular heading difference of at most $15^\circ$. Pairs with visual cosine similarity $V_{ij}\geq0.98$ are positive ($\mathcal{P}^{+}$), while those with $V_{ij}<0.1$ are negative ($\mathcal{P}^{-}$); intermediate pairs are ignored. Let $S_{ij}$ denote the cosine similarity between the corresponding path-integrated spatial representations.

\begin{equation}
\mathcal{L}_{\mathrm{pair}}
=\underset{(i,j)\in\mathcal{P}^{+}}{\mathrm{Avg}}
[0.9-S_{ij}]_{+}
+
\underset{(i,j)\in\mathcal{P}^{-}}{\mathrm{Avg}}
[S_{ij}-0.7]_{+},
\end{equation}

Here, $[x]_{+}=\max(x,0)$, and averages over empty sets are defined as zero. The loss encourages similar observations to have consistent spatial representations while separating visually dissimilar ones.

The student loss aligns the output of the action-only motion encoder $s_t^{act}$ with the output of the visual-action motion encoder $s_t^{obs}$ during training, using a $\operatorname{SmoothL1}$ loss with gradients stopped through the teacher, where \(\operatorname{sg}[\cdot]\) denotes the stop-gradient operation. The reconstruction and student losses are averaged over their respective valid time steps and batch elements.
\begin{equation}
\mathcal{L}_{\mathrm{student}}
=
\operatorname{SmoothL1}\!\left(
s_t^{\mathrm{act}},
\operatorname{sg}\!\left[s_t^{\mathrm{obs}}\right]
\right),
\end{equation}

\section{Experiments}
We collect trajectories from the $9\times9$ MemoryMaze and HM3D environments ~\citep{ramakrishnan2021hm3d} and evaluate open-loop visual prediction conditioned on actions and head directions. Each model
observes historical RGB frames, actions, and ground-truth head directions, and
autoregressively predicts future visual sequences from future actions and
ground-truth head directions. All models are trained on continuous 320-frame
trajectories using a $K=16$ frame open-loop prediction objective for
$30{,}000$ gradient steps on MemoryMaze and $100{,}000$ steps on HM3D; CoME instead retains its original diffusion-based training scheme. All models use a batch size of 16 and are trained
over three random seeds. Except for CoME (approximately 297M parameters), model sizes are controlled to 7–8M on MemoryMaze and approximately 100M on HM3D. We consider two continuous open-loop evaluation
protocols: predicting 64 frames from 256 context frames and predicting 500
frames from 500 context frames.

\begin{table*}[t]
\centering
\footnotesize
\setlength{\tabcolsep}{3pt}
\renewcommand{\arraystretch}{1.15}
\caption{Open-loop visual prediction on $9\times9$ MemoryMaze and HM3D. We report mean $\pm$ sample standard deviation over three training seeds. Best and second-best results are shown in bold and underlined, respectively, based on unrounded values.}
\label{tab:memorymaze-main}
\resizebox{\textwidth}{!}{%
\begin{tabular}{lcccccc}
\toprule
& \multicolumn{4}{c}{\textbf{MemoryMaze}} & \multicolumn{2}{c}{\textbf{HM3D}} \\
\cmidrule(lr){2-5}\cmidrule(lr){6-7}
Model & MSE $\downarrow$ & SSIM $\uparrow$ & LPIPS $\downarrow$ & PSNR $\uparrow$ & MSE $\downarrow$ & LPIPS $\downarrow$ \\
\midrule
\multicolumn{7}{l}{\textbf{256 context frames $\rightarrow$ 64 predicted frames}} \\
\addlinespace[1pt]
Transformer & $0.0100 \pm 0.0006$ & $0.7092 \pm 0.0094$ & $0.2886 \pm 0.0137$ & $22.02 \pm 0.47$ & $\underline{0.0137 \pm 0.0004}$ & $0.557 \pm 0.004$ \\
Mamba & $0.0171 \pm 0.0060$ & $0.6568 \pm 0.0348$ & $0.3873 \pm 0.0813$ & $19.01 \pm 2.14$ & $0.0340 \pm 0.0004$ & $0.739 \pm 0.003$ \\
RSSM & $0.0184 \pm 0.0011$ & $0.6269 \pm 0.0024$ & $0.3914 \pm 0.0035$ & $18.64 \pm 0.08$ & $0.0423 \pm 0.0012$ & $0.771 \pm 0.003$ \\
HPC--MEC & $\underline{0.0082 \pm 0.0001}$ & $\underline{0.7392 \pm 0.0015}$ & $0.2447 \pm 0.0030$ & $\underline{23.42 \pm 0.11}$ & $0.0137 \pm 0.0004$ & $0.573 \pm 0.004$ \\
CoME & $0.0171 \pm 0.0004$ & $0.5738 \pm 0.0017$ & $\underline{0.2345 \pm 0.0050}$ & $17.67 \pm 0.10$ & $0.0483 \pm 0.0035$ & $\underline{0.465 \pm 0.010}$ \\
\addlinespace[2pt]
HiSTFormer (ours) & $\boldsymbol{0.0059 \pm 0.0002}$ & $\boldsymbol{0.7639 \pm 0.0026}$ & $\boldsymbol{0.2124 \pm 0.0027}$ & $\boldsymbol{24.72 \pm 0.16}$ & $\boldsymbol{0.0115 \pm 0.0003}$ & $\boldsymbol{0.462 \pm 0.002}$ \\
\midrule
\multicolumn{7}{l}{\textbf{500 context frames $\rightarrow$ 500 predicted frames}} \\
\addlinespace[1pt]
Transformer & $0.0281 \pm 0.0022$ & $0.5931 \pm 0.0097$ & $0.4063 \pm 0.0069$ & $16.95 \pm 0.33$ & $0.0141 \pm 0.0013$ & $0.538 \pm 0.018$ \\
Mamba & $0.0244 \pm 0.0003$ & $0.6171 \pm 0.0027$ & $0.4354 \pm 0.0379$ & $17.02 \pm 0.38$ & $0.0491 \pm 0.0043$ & $0.786 \pm 0.003$ \\
RSSM & $0.0269 \pm 0.0038$ & $0.5752 \pm 0.0292$ & $0.4451 \pm 0.0160$ & $16.70 \pm 0.57$ & $0.0504 \pm 0.0054$ & $0.805 \pm 0.005$ \\
HPC--MEC & $\underline{0.0159 \pm 0.0004}$ & $\underline{0.6657 \pm 0.0012}$ & $0.3573 \pm 0.0064$ & $\underline{19.54 \pm 0.09}$ & $\underline{0.0115 \pm 0.0002}$ & $\underline{0.523 \pm 0.006}$ \\
CoME & $0.0263 \pm 0.0007$ & $0.4880 \pm 0.0059$ & $\underline{0.3169 \pm 0.0020}$ & $15.80 \pm 0.12$ & $0.0621 \pm 0.0045$ & $0.605 \pm 0.016$ \\
\addlinespace[2pt]
HiSTFormer (ours) & $\boldsymbol{0.0133 \pm 0.0005}$ & $\boldsymbol{0.7013 \pm 0.0047}$ & $\boldsymbol{0.2920 \pm 0.0037}$ & $\boldsymbol{21.15 \pm 0.17}$ & $\boldsymbol{0.0078 \pm 0.0004}$ & $\boldsymbol{0.401 \pm 0.009}$ \\
\bottomrule
\end{tabular}%
}
\end{table*}

\subsection{Baselines}
We compare HiSTFormer with recent world models equipped with spatial or long-term
memory mechanisms, as well as general-purpose sequence models with comparable
parameter counts. All baselines are adapted to a common observation, action,
and head-direction interface, while their original architectures and training
procedures are preserved whenever possible.

\noindent\textbf{HPC--MEC.}
We reimplement and adapt HPC--MEC \citep{zhang2026structure}, a hippocampal--entorhinal-inspired world model,
using the same head-direction-based coordinate transformation and data
interface as HiSTFormer. The resulting model contains approximately $8.15$ M
parameters.

\noindent\textbf{CoME.}
We adapt and train CoME on our dataset
following its official implementation \citep{stapf2026CoME}. To avoid
discrepancies arising from different pose-estimation implementations, we
provide ground-truth poses directly. CoME has substantially more parameters
than HiSTFormer. Appendix~\ref{app:extended_baselines} further reports results
under a longer training budget; the additional training improves performance
but does not change the main conclusions.

\noindent\textbf{General-purpose world models.}
We construct three latent-space world models based on Transformer, RSSM, and
Mamba \citep{vaswani2017attention, hafner2019learning, gu2024mamba, hafner2025dreamerv3}. They share the same visual encoder--decoder framework and input and
output interfaces. All three models are trained with the same open-loop
prediction and image-reconstruction losses, while retaining their
architecture-specific auxiliary latent-space objectives. Each model contains
approximately $8$ M parameters.

\section{Results}

\subsection{Long-Horizon Visual Prediction}

\begin{figure*}[t]
    \centering
    \includegraphics[width=\textwidth]{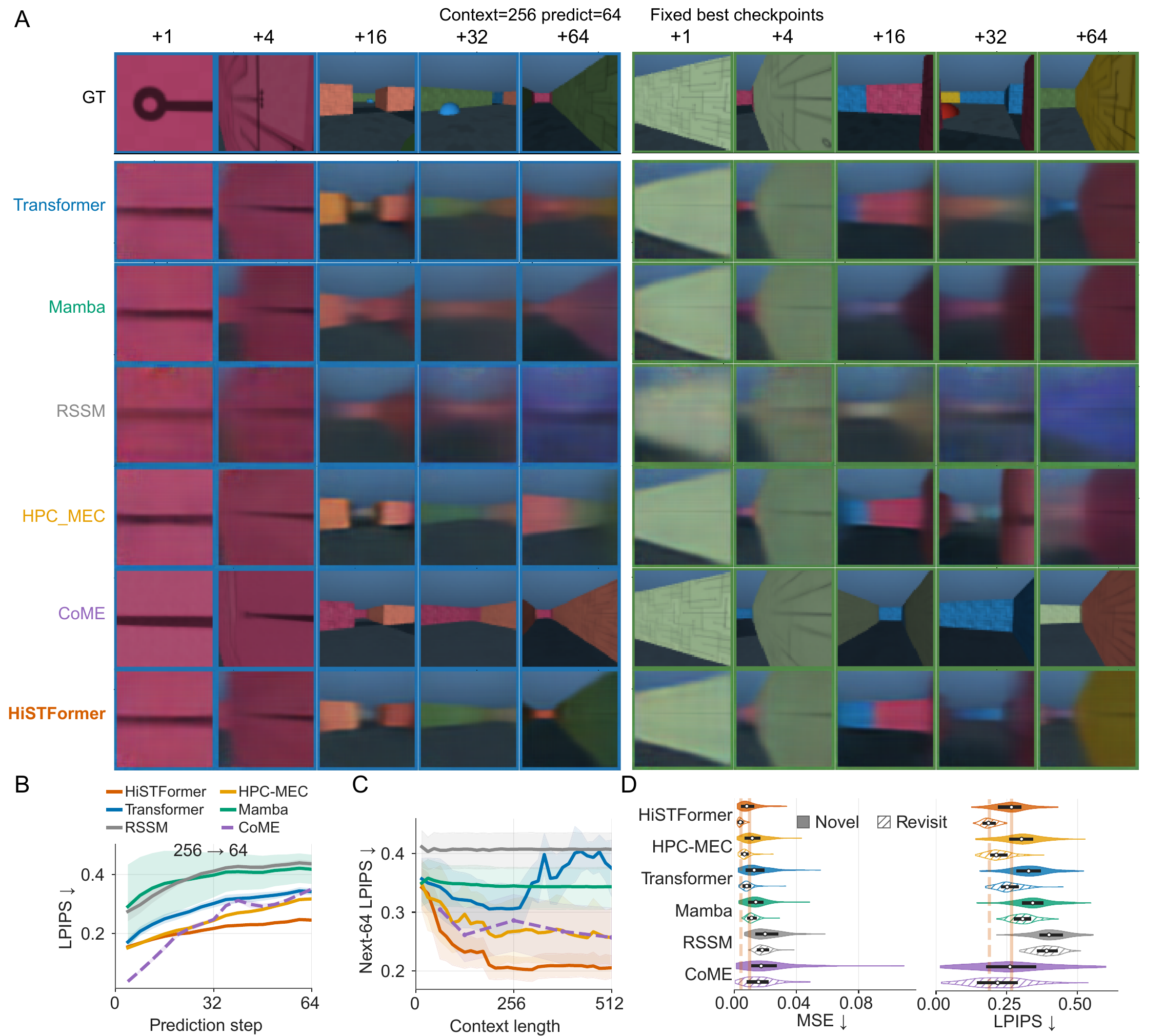}
    \caption{Analysis of long-horizon visual prediction in Memory Maze.
    \textbf{A}: Qualitative predictions at selected rollout horizons under the
    $256\!\rightarrow\!64$ protocol. \textbf{B}: Per-step LPIPS over the
    64-frame prediction horizon. \textbf{C}: LPIPS for the next 64 predicted
    frames as the available context length increases. \textbf{D}: MSE and
    LPIPS on spatially novel and revisited prediction frames.}
    \label{fig:visual_comparison}
\end{figure*}

Table~\ref{tab:memorymaze-main} compares visual prediction performance on MemoryMaze and HM3D under
the $256\!\rightarrow\!64$ and $500\!\rightarrow\!500$ protocols \citep{wang2004ssim,zhang2018lpips}. The
$256\!\rightarrow\!64$ protocol fits within the 320-frame training length but
uses a fourfold longer open-loop horizon. On MemoryMaze, HiSTFormer remains robust and reduces
LPIPS by $9.4\%$ relative to CoME.
The $500\!\rightarrow\!500$ protocol exceeds the training regime in both total
sequence length and prediction horizon. HiSTFormer again achieves the best overall
performance on MemoryMaze and widens its margin over most baselines, with HPC--MEC remaining
the closest competitor. CoME remains competitive only in LPIPS, whereas
HiSTFormer provides balanced gains in both pixel-level fidelity and perceptual
quality.

On HM3D, HiSTFormer also achieves the lowest MSE and LPIPS among the baselines in Table~\ref{tab:memorymaze-main} under both protocols, demonstrating consistent performance across diverse environments. Qualitative comparisons and further analysis are provided in the Appendix ~\ref{app:hm3d_adaptation}.

HiSTFormer's fixed-capacity structured memory keeps KV storage and per-step
attention cost bounded as the sequence grows. Appendix~\ref{app:memory_management}
further shows that bounding the spatial cache prevents KV growth with
negligible impact on prediction quality.

The qualitative comparison on MemoryMaze in Figure~\ref{fig:visual_comparison}A further
illustrates how prediction quality evolves with the rollout horizon. HiSTFormer
more consistently preserves scene layout, wall appearance, and viewpoint
evolution. CoME produces sharp images, but some long-horizon predictions
contain incorrect colors or scene geometry. Figure~\ref{fig:visual_comparison}B
reports per-step LPIPS under the $256\!\rightarrow\!64$ protocol. Beyond the
16-step open-loop horizon used during training, HiSTFormer exhibits the slowest
error accumulation, and its advantage increases with the rollout horizon.
Although CoME achieves low LPIPS during the first few prediction steps, its
error subsequently grows more rapidly.

To determine whether these long-horizon gains arise from exploiting additional
history, we fix the prediction target to the next 64 frames and progressively
increase the available context length
(Figure~\ref{fig:visual_comparison}C). HiSTFormer's error decreases as the context
grows and then remains stable when the context extends beyond the training
sequence length, despite accumulated path-integration error and a
fixed-capacity KV cache. This behavior indicates a stronger ability to extract
useful information from long histories. HPC--MEC and CoME also benefit from additional context,
whereas Mamba and RSSM show only limited improvement. Transformer benefits from
longer context within the training range, but its performance degrades when the
context is extended further.

Finally, we separately evaluate predictions at spatially revisited and
spatially novel locations (Figure~\ref{fig:visual_comparison}D). A target is
revisited if a context observation occurred at least 32 frames earlier within
$2$~m of its position and with a heading difference of at most $45^\circ$; it
is novel if every context observation is more than $2$~m away. Other targets
are excluded. HiSTFormer achieves the lowest MSE on both subsets and outperforms
most baselines in LPIPS, although its LPIPS difference from CoME on novel
targets is not statistically significant. These results demonstrate robust
prediction at both revisited and novel locations under a constrained memory
budget.

\subsection{Hierarchical Temporal--Spatial Encoding Enables Scalable
Long-Horizon Prediction}
\label{sec:temporal_spatial_results}
Local temporal continuity and long-range spatial relationships exhibit
different structures, motivating us to examine whether these two types of
positional information should be organized differently within a Transformer.
We systematically compare temporal RoPE (T), spatial RoPE (S), within-layer
joint encoding (J) \citep{wang2024qwen2}, no positional encoding (N), and different layer-wise
temporal--spatial arrangements, evaluating both in-range performance and
long-horizon generalization.

All configurations use the same 128-token memory budget. Temporal layers use a
contiguous FIFO window, whereas spatial layers use recent-plus-coverage (R+C):
32 recent tokens and 96 earlier memories retained for temporal coverage.
Unless cache organization is under study, each layer uses its default policy;
content-conditioned spatial offsets are disabled to isolate positional
encoding and layer-wise organization. We report LPIPS under the
$256\!\rightarrow\!64$ and $500\!\rightarrow\!500$ protocols to assess
long-horizon generalization within and beyond the training sequence length.

\begin{figure*}[t]
    \centering
    \includegraphics[width=\textwidth]{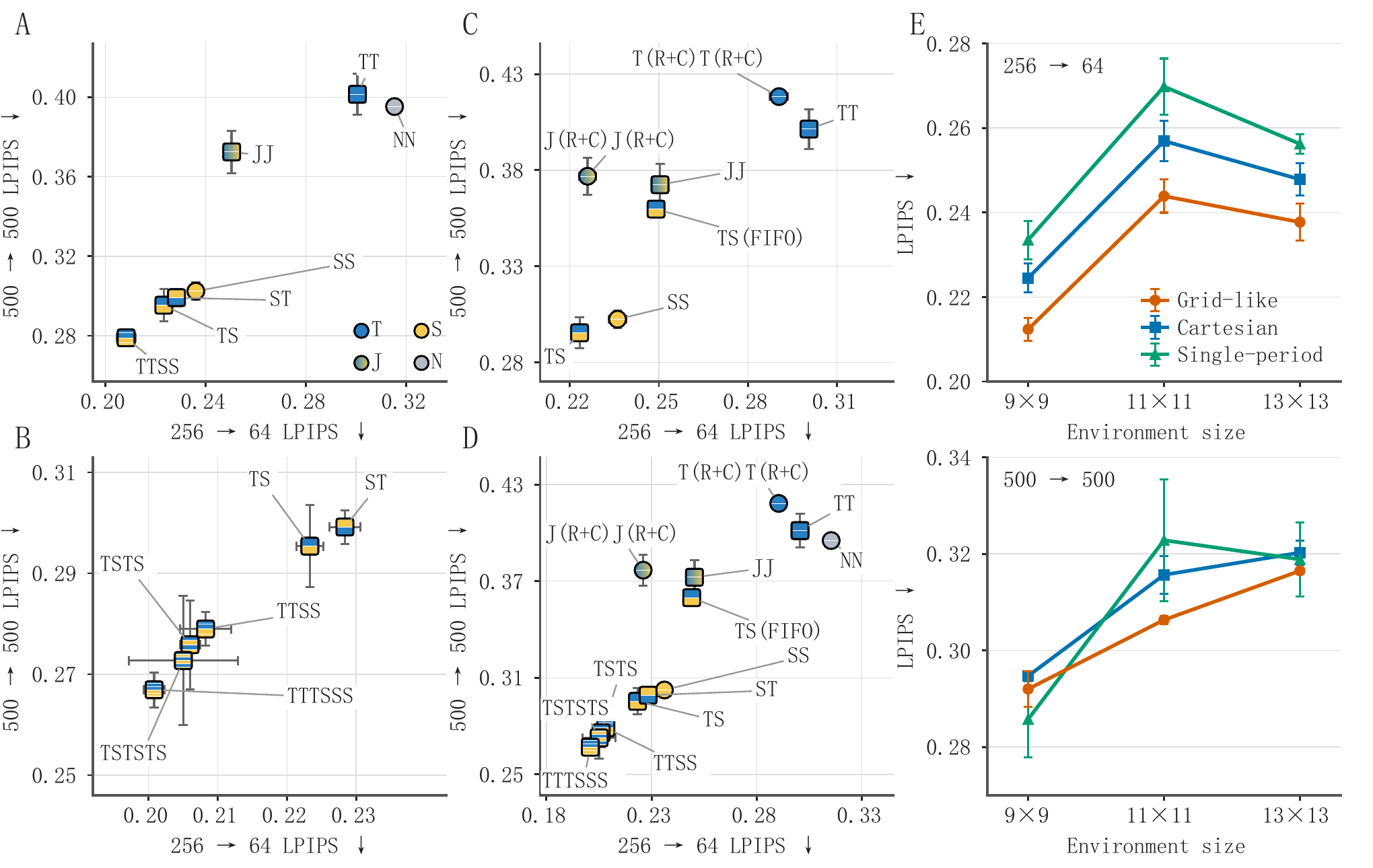}
    \caption{Analysis of hierarchical temporal--spatial encoding.
    \textbf{A}: Comparison of positional encodings in two-layer models. \textbf{B}: Temporal--spatial layer arrangements at depths
    of 2, 4, and 6 layers. \textbf{C}: Effects of exchanging the default
    cache organizations of temporal and spatial layers. \textbf{D}: Combined
    comparison of positional encoding, hierarchy, and cache-organization
    variants. \textbf{E}: Zero-shot generalization of different spatial
    encodings from the $9\times9$ training environment to larger environments
    under the $256\!\rightarrow\!64$ and $500\!\rightarrow\!500$ protocols.}
    \label{fig:temporal_spatial_analysis}
\end{figure*}

\paragraph{Temporal and spatial positional structures are complementary.}
Figure~\ref{fig:temporal_spatial_analysis}A compares the main encoding schemes
at the same two-layer depth. TT, which uses only temporal RoPE, and NN, which
uses no positional encoding, both perform poorly on long-horizon prediction.
In contrast, SS, which uses only spatial RoPE, substantially reduces the error
under the $500\!\rightarrow\!500$ protocol, highlighting the importance of
explicit spatial indexing for long-range prediction. JJ, which directly
combines temporal and spatial signals within each layer, fails to realize the
full benefit of both and performs between the purely temporal and purely
spatial variants. When temporal and spatial relations are instead assigned to
different Transformer layers, both TS and ST achieve stronger overall
performance, with TS---temporal contextualization followed by spatial
organization---performing better. These results indicate that temporal and
spatial positional structures provide complementary information, but the
benefit depends on how they are organized within the network; simple
within-layer mixing is insufficient to fully exploit their complementarity.

\paragraph{Hierarchical organization consistently benefits from increased
depth.}
We further compare temporal--spatial Transformers with different depths and
layer arrangements (Figure~\ref{fig:temporal_spatial_analysis}B). As the
network is scaled from 2 to 4 and 6 layers, LPIPS consistently decreases
under both the $256\!\rightarrow\!64$ and $500\!\rightarrow\!500$ protocols,
indicating that the hierarchical temporal--spatial design continues to benefit
from increased model depth. At six layers, the blockwise TTTSSS arrangement
achieves lower prediction error than the alternating TSTSTS arrangement. At
both four and six layers, the blockwise arrangements (TTSS and TTTSSS) also
exhibit smaller variation across random seeds than their alternating
counterparts (TSTS and TSTSTS). Within the
tested range of two to six layers, these results show that placing local
temporal processing before long-range spatial organization provides stable
depth scaling.

\paragraph{Temporal and spatial layers benefit from different memory
organizations.}
In the experiments above, temporal layers access a contiguous window of recent
history by default, whereas spatial layers combine a recent window with
long-term memories selected to provide broad temporal coverage. To examine the
effect of this asymmetric design, we further swap the cache organization used
by the two layer types (Figure~\ref{fig:temporal_spatial_analysis}C). Applying
R+C coverage sampling to temporal layers does not yield consistent improvements
and leads to a clear degradation in long-horizon prediction. Likewise,
replacing the coverage-based memory of spatial layers with a pure FIFO cache
degrades the performance of TS. In contrast, the default
configuration---contiguous recent history for temporal layers and long-range
coverage for spatial layers---achieves the best overall performance. This
asymmetry suggests that, under a limited memory budget, local temporal modeling
relies more strongly on contiguous context, whereas spatial retrieval benefits
more from broad coverage over long histories.
Figure~\ref{fig:temporal_spatial_analysis}D summarizes all variants discussed
above for a complete comparison.

Taken together with the spatial-cache organization experiments in
Appendix~\ref{app:memory_management}, these results show that spatial indexing
allows the model to exploit information spanning long temporal ranges without
retaining the full history. Consequently, KV storage and per-step attention
cost remain bounded as the sequence grows.

\paragraph{Spatial modeling is robust across encoding choices.}
Keeping all other settings unchanged, we compare the multi-periodic
grid encoding with two alternatives: Cartesian, which uses orthogonal wave directions, 
and Single-period, which retains the grid-like wave directions but uses a single period
 (Figure~\ref{fig:temporal_spatial_analysis}E). We
evaluate zero-shot transfer from the $9\times9$ training environment to larger
$11\times11$ and $13\times13$ environments, averaging results over three
training seeds. Under the $256\!\rightarrow\!64$ protocol, the multi-periodic
encoding achieves the lowest LPIPS across all environment sizes.
Under the more challenging $500\!\rightarrow\!500$ setting, the differences
among encodings narrow, while the multi-periodic representation remains stable
and competitive across environment scales. 
Overall, these results suggest that HiSTFormer's spatial modeling is robust to the tested encoding choices, while the multi-periodic grid-like representation provides a reliable default. Multi-scale spatial codes have theoretical support for reducing periodic positional ambiguity~\citep{mathis2012optimal,stemmler2015connecting}. Further implementation details and distance-stratified analyses are provided in Appendix~\ref{app:spatial_encoding_design}.

\subsection{Learnable Address Offsets Improve Spatial Organization}
\label{sec:address_offset_results}
Temporal layers aggregate visual information across frames, so their
contextual representations may not correspond to a single agent-centered
position from path integration. HiSTFormer therefore predicts a
content-conditioned two-dimensional offset from the temporal output for each
spatial-attention head, adapting the base path-integrated address to the
represented content.

\begin{figure}[!ht]
    \centering
    \includegraphics[width=\textwidth]{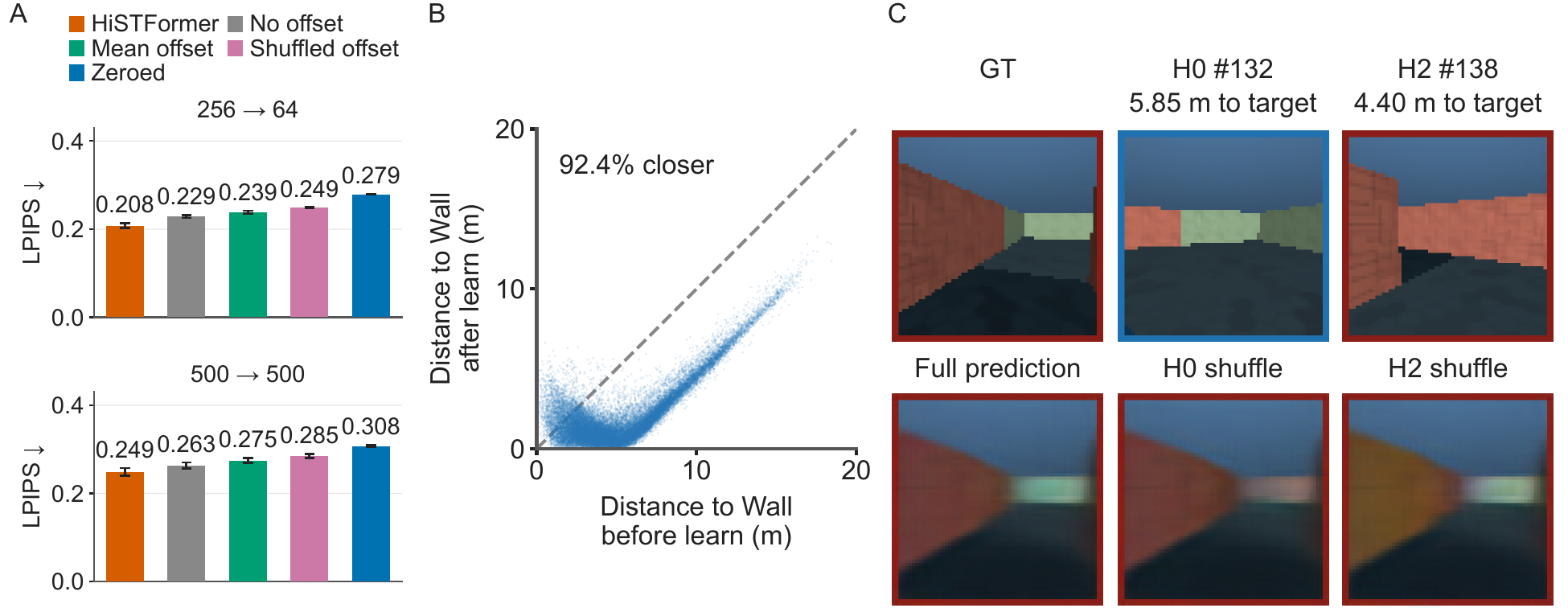}
    \caption{Analysis of content-conditioned learnable address offsets.
    \textbf{A}: LPIPS of the full HiSTFormer, a model trained without address
    offsets, and three test-time interventions applied after normal training:
    replacing the learned offsets with their mean, shuffling them, or setting
    them to zero. \textbf{B}: Distance from the base and offset-adjusted spatial
    addresses to the wall intersected by the current viewing direction.
    \textbf{C}: A qualitative example showing the ground-truth target,
    high-attention memories retrieved by spatial heads H0 and H2, the full
    prediction, and predictions obtained after shuffling the retrieval results
    of either head.}
    \label{fig:claim3_combined}
\end{figure}

\paragraph{Learnable address offsets consistently improve visual prediction.}
We compare the full HiSTFormer with learned address offsets, a No-offset variant
trained from scratch with offsets disabled, and three variants obtained by
modifying the learned offsets only at test time after normal training: taking
their mean, shuffling them, or setting them to zero
(Figure~\ref{fig:claim3_combined}A). The full HiSTFormer achieves the lowest
LPIPS under both the $256\!\rightarrow\!64$ and $500\!\rightarrow\!500$
protocols, reducing LPIPS by approximately $9.0\%$ and $5.5\%$, respectively,
relative to the No-offset model. All three test-time interventions degrade
performance. These results show that
content-conditioned offsets improve the learned representations and actively
adjust spatial addresses during inference.

\paragraph{The learned address offsets align with visible scene geometry.}
We measure the distance from both the path-integrated base address and the
head-specific adjusted address to the wall intersected by the current viewing
direction (Figure~\ref{fig:claim3_combined}B). After applying the learned
offset, $92.4\%$ of samples move closer to the corresponding wall location,
despite receiving no direct supervision of wall positions. Together with the
performance degradation under offset interventions in
Figure~\ref{fig:claim3_combined}A, this result indicates that the offsets are
structured, content-dependent spatial adjustments rather than arbitrary
perturbations.
Further head-wise analysis (Appendix Figure~\ref{app:learn_offset} A) reveals relatively stable directional preferences, with the four heads distributed approximately uniformly across the 80° field of view and each head's directions concentrated within a 20° range. Offset magnitudes are positively correlated with the distance to the nearest wall along the corresponding direction, most strongly within 3 environment cells, before gradually saturating. Representative offset patterns are shown in Appendix Figure~\ref{app:learn_offset} B.

\paragraph{Adjusted spatial addresses retrieve historical visual cues relevant
to the current prediction.}
Finally, we examine how these address adjustments contribute to visual
prediction in practice. To reduce the effect of directly retrieving historical
views from the same location, we select target frames whose positions are more
than $4$~m away from all ground-truth context observations.
Figure~\ref{fig:claim3_combined}C shows a representative example.
High-attention memory entries retrieved by spatial heads H0 and H2 are located
$5.85$~m and $4.40$~m from the target position, respectively, and contain
partial information about the target wall observed from different viewpoints,
while neither contains the complete target view. The full model reconstructs
the wall structure and color of the target scene reasonably well, whereas
shuffling the retrieval results of either H0 or H2 leads to clear prediction
errors. This example suggests that content-conditioned addressing helps the
model retrieve prediction-relevant information from spatially distinct
historical experiences that provide complementary visual cues, and incorporate
these cues into subsequent visual generation. Appendix Figure~\ref{app:learn_offset} C/D/E further analyzes the head-specific attended content and memory perturbations in this case, yielding consistent findings.

Overall, these results indicate that the learnable address offsets do more than
simply introduce additional degrees of freedom. Instead, they provide
content-dependent spatial adjustments to the base positions obtained through
path integration. These adjustments yield consistent prediction gains,
exhibit interpretable structure aligned with scene geometry, and influence the
historical memories actually used by the model, thereby enabling a more
flexible binding between temporal context and spatial memory.

\section{Conclusion}
We introduced HiSTFormer, a hierarchical Transformer that 
organizes local temporal integration and long-range spatial retrieval into successive stages, with content-conditioned offsets for adaptive memory addressing. Using a fixed-capacity memory and grid-like spatial encoding, HiSTFormer improves long-horizon visual prediction both within and beyond the training sequence length.

\section{Limitations}
HiSTFormer assumes reliable ground-truth head directions. Its robustness to noisy heading estimates and generalization to downstream tasks remain to be explored.

\clearpage

\subsection*{AI use statement}
We used generative AI tools to assist with research ideation, literature discovery, drafting of some paper sections, and language refinement. AI tools were used to help explore research directions, identify related work, and improve the clarity and presentation of the manuscript. All AI-assisted content, including generated text and suggestions, was carefully reviewed and verified by the authors. The authors take full responsibility for the final content, claims, experimental results, and conclusions presented in this paper.

\subsection*{Reproducibility statement}

We provide implementation details of HiSTFormer, training configurations, and
evaluation protocols in the appendix. We also provide detailed
descriptions of the dataset construction, baseline implementations, and
experimental settings to facilitate reproduction of our results.

\bibliography{iclr2027_conference}
\bibliographystyle{iclr2027_conference}

\appendix
\section{Implementation and Training Details}
\label{app:implementation_details}

\subsection{Visual Encoder and Decoder}
HiSTFormer uses a convolutional encoder--decoder. The encoder takes a
$64\times64\times3$ RGB image as input and consists of four downsampling
convolutional blocks. Each block contains a $3\times3$ convolution with stride
2, followed by GroupNorm and SiLU. The channel dimensions are 32, 64, 128, and
128, respectively.

The resulting $4\times4\times128$ feature map is flattened and passed through
a multilayer perceptron (MLP), producing an $Z_D=234$-dimensional visual
representation:
\begin{equation}
\mathbf{z}_t = E_{\phi}(o_t).
\end{equation}
The decoder first maps the latent representation back to a
$4\times4\times128$ feature map through a linear layer, followed by four stages
of nearest-neighbor upsampling and convolution to recover a $64\times64$ RGB
image. A sigmoid activation is applied at the output:
\begin{equation}
\hat{o}_{t+1}
=
D_{\psi}\!\left(\hat{\mathbf{z}}_{t+1}\right).
\end{equation}
The encoder, predictive model, and decoder are trained jointly from scratch.

\subsection{Motion Encoder and Teacher--Student Alignment}
\label{app:motion_encoder}
The motion encoder uses a causal temporal convolution with kernel size 9 and
dilation 1, covering the current input and the preceding eight steps. Given the
action $\mathbf{a}_t$ and, when available, the visual representation $z_t$, the observation branch predicts the displacement
$s_t^{\mathrm{obs}}$. During open-loop prediction, an action-only branch
predicts $s_t^{\mathrm{act}}$.

The observation branch acts as the teacher and the action-only branch as the
student. Their outputs are aligned using a smooth $L_1$ loss, with gradients
stopped through the teacher:
\begin{equation}
\mathcal{L}_{\mathrm{student}}
=
\operatorname{SmoothL1}\!\left(
s_t^{\mathrm{act}},
\operatorname{sg}\!\left[s_t^{\mathrm{obs}}\right]
\right),
\end{equation}
where $\operatorname{sg}[\cdot]$ denotes the stop-gradient operation.

\subsection{Main Model Configuration}
\label{app:main_model_configuration}
The main HiSTFormer model on MemoryMaze reported in Table~\ref{tab:memorymaze-main} contains
approximately 7.72 million trainable parameters and uses a two-layer TS
Transformer, with one temporal layer followed by one spatial layer. The
temporal layer maintains a 32-entry FIFO KV cache.

For each spatial-attention head, the spatial layer applies RoPE to 128 of the
256 query and key dimensions. Positional RoPE occupies 96 dimensions, using
eight spatial periods with six rotary channel pairs per period. Head direction
occupies a further 32 dimensions by repeating the corresponding rotation
across 16 channel pairs. The remaining 128 dimensions are not rotated.

The spatial layer receives the complete output of the temporal layer. For each
token, a content-conditioned address network with dimensions
$384\!\rightarrow\!128\!\rightarrow\!8$ produces eight scalars(0.05M), which are
reshaped into four two-dimensional position offsets, one for each attention
head. These offsets adjust the spatial rotary phases used by the corresponding
heads.

\subsection{Long-Term Memory Management}
\label{app:memory_management}
By default, HiSTFormer manages its fixed-capacity long-term memory using
temporal-redundancy-based eviction. Let the retained entries be ordered by
their timestamps. When the long-term memory is full, we assign each interior
candidate at timestamp $\tau_j$ the temporal-redundancy score
\begin{equation}
C_j
=
(\tau_j-\tau_{j-1})
(\tau_{j+1}-\tau_j)
(\tau_{j+1}-\tau_{j-1}).
\end{equation}
The entry with the smallest $C_j$ is removed, while the two temporal endpoints
are always retained. This strategy preferentially removes temporally redundant
entries and maintains coverage over a broad temporal range.

\paragraph{Sensitivity to the memory-management strategy.}
To assess whether HiSTFormer depends on a particular eviction heuristic, we keep
the checkpoints trained with the default temporal-coverage strategy fixed and
modify only the memory-management policy used at test time. The fixed-interval
strategy retains one long-term entry every $k$ frames, with $k=4$ for the
$256\!\rightarrow\!64$ protocol and $k=8$ for the
$500\!\rightarrow\!500$ protocol. The random strategy discards $80\%$ of the
candidate entries. As shown in Table~\ref{tab:memory_management}, prediction
quality remains stable despite substantial differences in the number and
distribution of retained memories. Relative to the default strategy, MSE
changes by at most $4.3\%$ under the $256\!\rightarrow\!64$ protocol and
$0.9\%$ under the $500\!\rightarrow\!500$ protocol. These results indicate
that HiSTFormer can flexibly exploit its spatial memory and is not strongly tied to
a specific eviction rule.

\begin{table}[!ht]
\centering
\small
\setlength{\tabcolsep}{3.5pt}
\renewcommand{\arraystretch}{1.12}
\caption{Sensitivity to the test-time long-term memory-management strategy.
All results use checkpoints trained with the default temporal-coverage
strategy. The number of long-term entries is averaged over evaluation
sequences.}
\label{tab:memory_management}
\resizebox{\textwidth}{!}{%
\begin{tabular}{llrrrrr}
\toprule
Protocol & Strategy & Long-term entries & MSE $\downarrow$ & SSIM $\uparrow$
& LPIPS $\downarrow$ & PSNR $\uparrow$ \\
\midrule
$256\!\rightarrow\!64$ & Temporal coverage (default)
& 96 & 0.006216 & 0.7587 & 0.2205 & 24.400 \\
& Fixed interval ($k=4$)
& 56 & 0.006142 & 0.7582 & 0.2218 & 24.323 \\
& Random eviction ($80\%$)
& 43.56 & 0.006480 & 0.7531 & 0.2294 & 24.095 \\
\midrule
$500\!\rightarrow\!500$ & Temporal coverage (default)
& 96 & 0.013600 & 0.6989 & 0.2926 & 21.166 \\
& Fixed interval ($k=8$)
& 59 & 0.013712 & 0.6961 & 0.2973 & 21.037 \\
& Random eviction ($80\%$)
& 89.58 & 0.013623 & 0.6985 & 0.2945 & 21.100 \\
\bottomrule
\end{tabular}
}
\end{table}

\paragraph{Bounded versus full spatial memory.}
Beyond changing the cache policy only at test time, we also conduct a
from-scratch comparison using the SS model. The full-memory variant uses the
complete spatial KV cache during both training and inference, whereas the
bounded variant uses recent-plus-coverage (R+C) memory with at most 128 cached
entries in both stages. As shown in
Table~\ref{tab:full_vs_bounded_spatial_memory}, the four image-prediction
metrics exhibit negligible degradation overall, with several metrics showing
modest improvements. R+C also keeps the peak spatial KV cache at 1.000~MiB
under both protocols, whereas the full cache grows from 2.492 to 7.805~MiB as
the sequence length increases. Thus, explicit spatial memory management bounds
KV storage without sacrificing prediction quality, highlighting its empirical
importance for long-horizon prediction. We further hypothesize that such sparse
memory may help mitigate memory interference caused by accumulated
path-integration errors, although directly validating this effect remains
future work.

\begin{table}[!ht]
\centering
\small
\setlength{\tabcolsep}{4pt}
\renewcommand{\arraystretch}{1.12}
\caption{From-scratch comparison of full and bounded spatial KV caches using
the SS model. R+C retains at most 128 entries during both training and
inference. Peak spatial KV-cache memory is reported in MiB.}
\label{tab:full_vs_bounded_spatial_memory}
\resizebox{\textwidth}{!}{%
\begin{tabular}{llrrrrr}
\toprule
Protocol & Training / inference cache & Peak KV (MiB) $\downarrow$
& MSE $\downarrow$ & SSIM $\uparrow$ & LPIPS $\downarrow$ & PSNR $\uparrow$ \\
\midrule
$256\!\rightarrow\!64$ & Full / Full
& 2.492 & 0.006856 & 0.747846 & 0.234693 & 23.844 \\
& R+C / R+C
& 1.000 & 0.006602 & 0.749573 & 0.231666 & 23.922 \\
\midrule
$500\!\rightarrow\!500$ & Full / Full
& 7.805 & 0.014121 & 0.689286 & 0.307988 & 20.648 \\
& R+C / R+C
& 1.000 & 0.014261 & 0.689550 & 0.304276 & 20.582 \\
\bottomrule
\end{tabular}
}
\end{table}

\section{Dataset Details}
\label{app:dataset_details}
\subsection{MemoryMaze dataset details}
\begin{figure}[t]
    \centering
    \includegraphics[width=\textwidth]{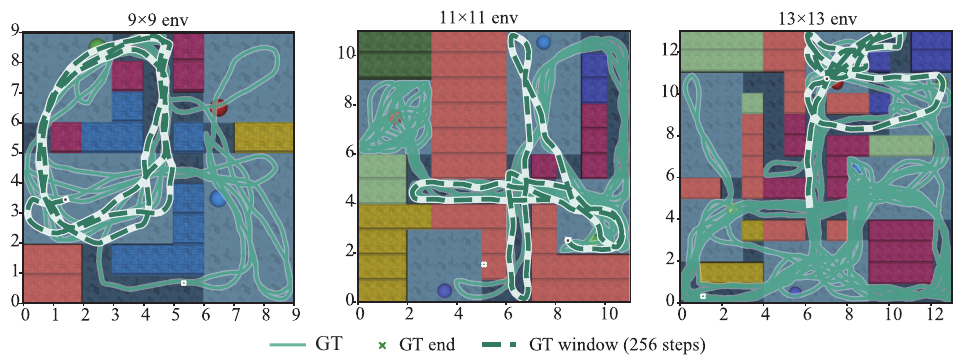}
    \caption{Representative layouts and sampled trajectories in
    $9\times9$, $11\times11$, and $13\times13$ Memory Maze environments. The
    dashed segment marks a 256-step ground-truth window, and the cross marks
    its endpoint.}
    \label{fig:dataset_overview}
\end{figure}

We collect $2{,}368$ training episodes and 112 test episodes in Memory Maze.
Each episode is generated from an environment initialized with a different
random seed and contains $1{,}000$ frames. Figure~\ref{fig:dataset_overview}
shows representative layouts and sampled trajectories for the $9\times9$,
$11\times11$, and $13\times13$ Memory Maze environments.

\begin{figure}[t]
    \centering
    \includegraphics[width=0.62\textwidth]{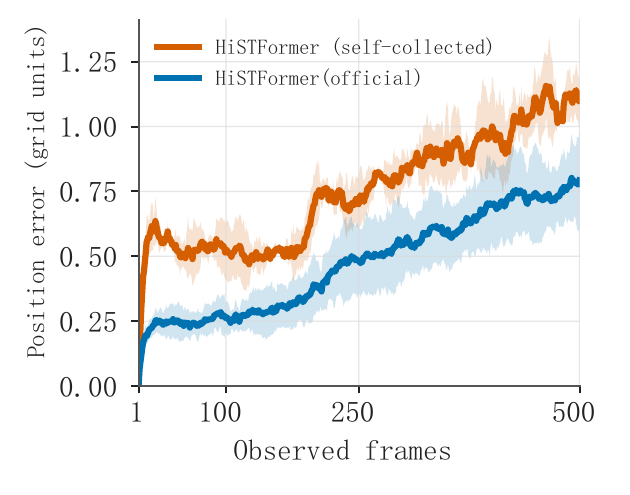}
    \caption{Accumulation of path-integration error in the main HiSTFormer model
    on our collected and the official datasets. Position is decoded from the
    largest-period component of the grid-like representation and compared with
    the ground-truth position over the first 500 observed frames. The larger
    error on our collected trajectories indicates more severe localization
    drift and a more challenging setting for long-horizon visual prediction.}
    \label{fig:position_drift}
\end{figure}

In the training set, $4.4\%$ of actions result in wall collisions, leading to
relatively large discrepancies between the intended action-induced motion and
the actual displacement. Consequently, path-integration errors are difficult
to avoid over long trajectories, testing the model's ability to integrate
spatial memories and resolve ambiguity as the underlying spatial
representation drifts.

Figure~\ref{fig:position_drift} compares the path-integration error decoded
from the main HiSTFormer model on our collected dataset and the official dataset.
For both datasets, we decode the estimated position from the largest-period
component of the grid-like representation and measure its deviation from the
ground-truth position over the first 500 observed frames. Although the error
accumulates with context length in both settings, the drift is substantially
more pronounced on our collected trajectories, making long-horizon visual
prediction more challenging. The multi-periodic representation,
content-conditioned offsets, and visual content inputs provide complementary
mechanisms that may partially mitigate the resulting performance loss by
helping disambiguate drifting spatial states, although they do not eliminate
the underlying localization error.

The trajectories also contain frequent revisits: for
$70.98\%$ of frames, the agent had previously visited a location within
$1$~m of its current position more than 32 frames earlier. This setting
therefore places stronger demands on how models organize and retrieve spatial
information.

Unless otherwise specified, we report experiments on our collected
$9\times9$ Memory Maze dataset. All experiments in
Section~\ref{sec:address_offset_results} instead use the official Memory Maze
dataset. Its less severe path-integration drift provides a cleaner setting for
isolating the effect of the learnable address offsets and also results in lower
overall prediction error.

\subsection{HM3D dataset details}
We collect an egocentric video dataset in HM3D v0.2 using Habitat-Sim. Our exploration controller, adapted from EgoEnv, initializes each episode at a sampled navigable location with a random heading and navigates among up to eight nearby reachable goals, reshuffling the goals to induce repeated visits. Each trajectory contains 1,024 consecutive observations, with RGB images rendered at $128\times 128$ and downsampled to $64 \times 64$. The action space consists of moving forward by 0.25 m and turning left or right by 30°. We record the executed actions together with ground-truth agent positions, camera poses, and headings. The final dataset contains 8,080 trajectories and 8,273,920 frames across 888 scenes. We use mutually disjoint sets of 788, 90, and 10 scenes for training, validation, and testing, comprising 7,880, 180, and 20 trajectories, respectively. Training scenes come from the HM3D training split, excluding 12 scenes with insufficient reachable goals. We reserve HM3D minival for our internal test set and remove these scenes from the validation split. All trajectories undergo geometric consistency checks and image/action replay verification at four checkpoints per trajectory.
\begin{figure*}[t]
    \centering
    \includegraphics[width=\textwidth]{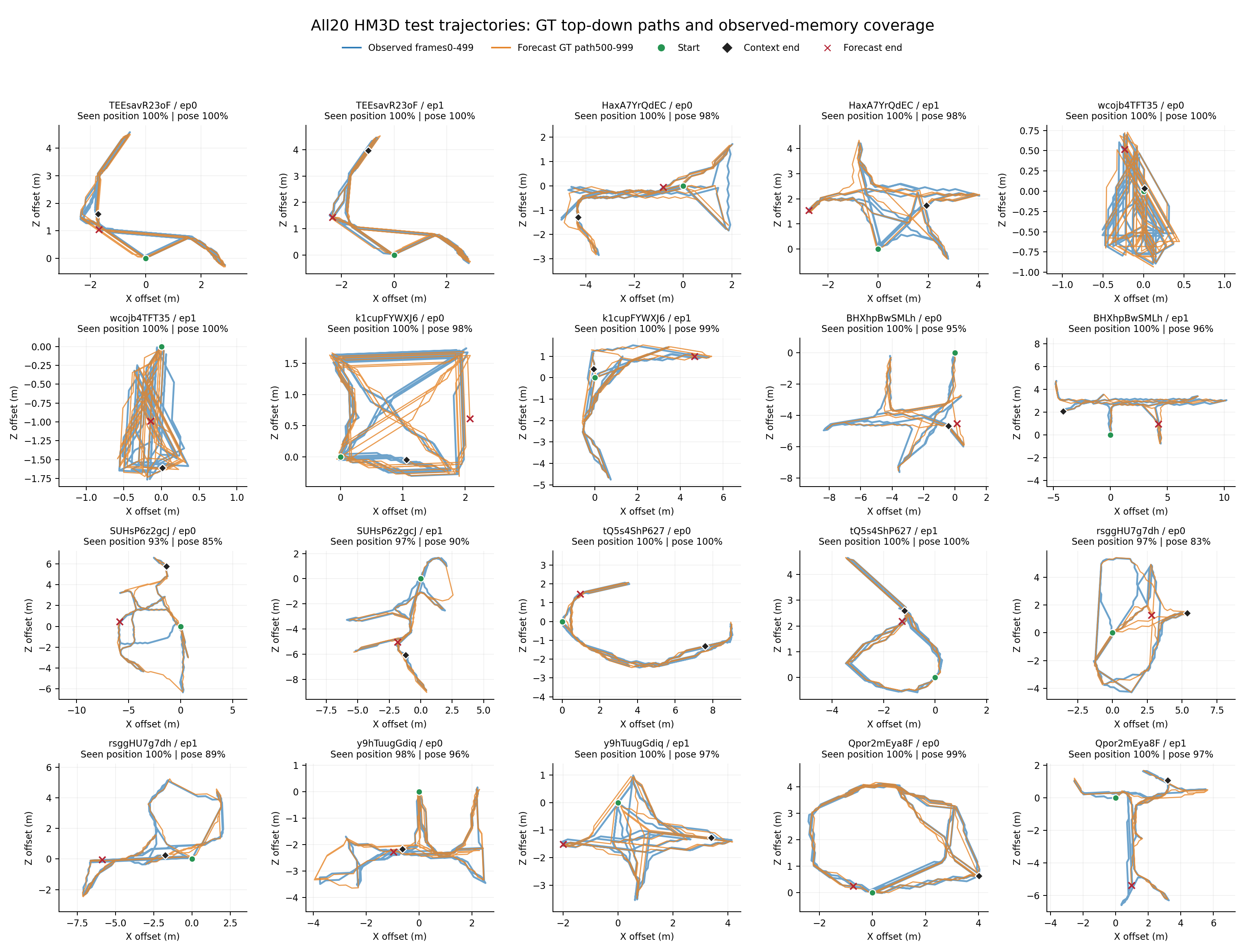}
    \caption{HM3D test trajectories and historical observation coverage. Twenty ground-truth trajectories from 10 test scenes are shown under the $500\!\rightarrow\!500$ protocol. Blue and orange lines indicate the observed 500-frame trajectories and the subsequent 500-frame ground-truth trajectories, respectively. Green circles, black diamonds, and red crosses mark the start, observation endpoint, and prediction endpoint. Each panel reports the position and pose matching rates against observed history, using thresholds of 0.5 m and $30^\circ$.}
    \label{fig:hm3d_traj}
\end{figure*}

\textbf{High Spatial Revisit Rates in HM3D.}
HM3D trajectories exhibit substantial spatial revisits, as agents sequentially navigate to eight random goals. Figure~\ref{fig:hm3d_traj} illustrates 20 test trajectories from 10 scenes and their historical observation coverage. We measure whether each predicted state matches an observed position (distance $\leq 0.5$), position and heading (angular difference $\leq 30^\circ$), or state-action transition (matching start/end states and action). Only observed history is considered, excluding predicted states.

Table~\ref{tab:revisit_coverage} reports the matching rates on both datasets. On HM3D, the three rates reach 87.27\%, 71.95\%, and 65.39\% under the $256\!\rightarrow\!64$ protocol, increasing to 99.27\%, 96.06\%, and 92.32\% under $500\!\rightarrow\!500$, all higher than on MemoryMaze. The previously reported 70.98\% position revisit rate on MemoryMaze was computed over 1,000 frames and thus differs slightly from the table. The two datasets also use different spatial units: maze coordinates for MemoryMaze and meters for HM3D.

\begin{table*}[t]
\centering
\small
\setlength{\tabcolsep}{8pt}
\renewcommand{\arraystretch}{1.15}
\caption{Historical observation coverage on MemoryMaze and HM3D. We report the percentage of predicted frames matching an observed position (distance $\leq 0.5$), position and heading (angular difference $\leq 30^\circ$), or state-action transition (matching start/end states and action). Only observed history is used for matching. Distance is measured in maze coordinate units for MemoryMaze and meters for HM3D.}
\label{tab:revisit_coverage}
\begin{tabular}{lcccc}
\toprule
& \multicolumn{2}{c}{$256\!\rightarrow\!64$} & \multicolumn{2}{c}{$500\!\rightarrow\!500$} \\
\cmidrule(lr){2-3}\cmidrule(lr){4-5}
Matching criterion & MemoryMaze & HM3D & MemoryMaze & HM3D \\
\midrule
Position & 69.27\% & 87.27\% & 84.78\% & \textbf{99.27\%} \\
Position and heading & 53.27\% & 71.95\% & 71.99\% & 96.06\% \\
State-action transition & 42.09\% & 65.39\% & 59.43\% & 92.32\% \\
\bottomrule
\end{tabular}
\end{table*}

\section{Adaptation to Complex Indoor Environments}
\label{app:hm3d_adaptation}
To assess HiSTFormer's adaptability to more complex indoor environments, we
conduct additional experiments on a first-person trajectory dataset built
from HM3D. Visual observations are represented using the pretrained Stable
Diffusion VAE~\citep{rombach2022highresolution} \texttt{sd-vae-ft-mse}. We freeze its encoder and decoder and
train only the latent-space prediction model. The HiSTFormer model uses a TTSS layer
arrangement with eight attention heads and a head dimension of 512, resulting
in 106.49 million trainable parameters; the spatial RoPE design remains
unchanged. Ground-truth $x, y, z$ positions and headings are provided during both
observation and prediction, while future RGB frames are never given to the
predictor. Training uses 16-step autoregressive rollouts with latent-space MSE
and runs for $100{,}000$ gradient updates.

\begin{figure*}[t]
    \centering
    \includegraphics[width=\textwidth]{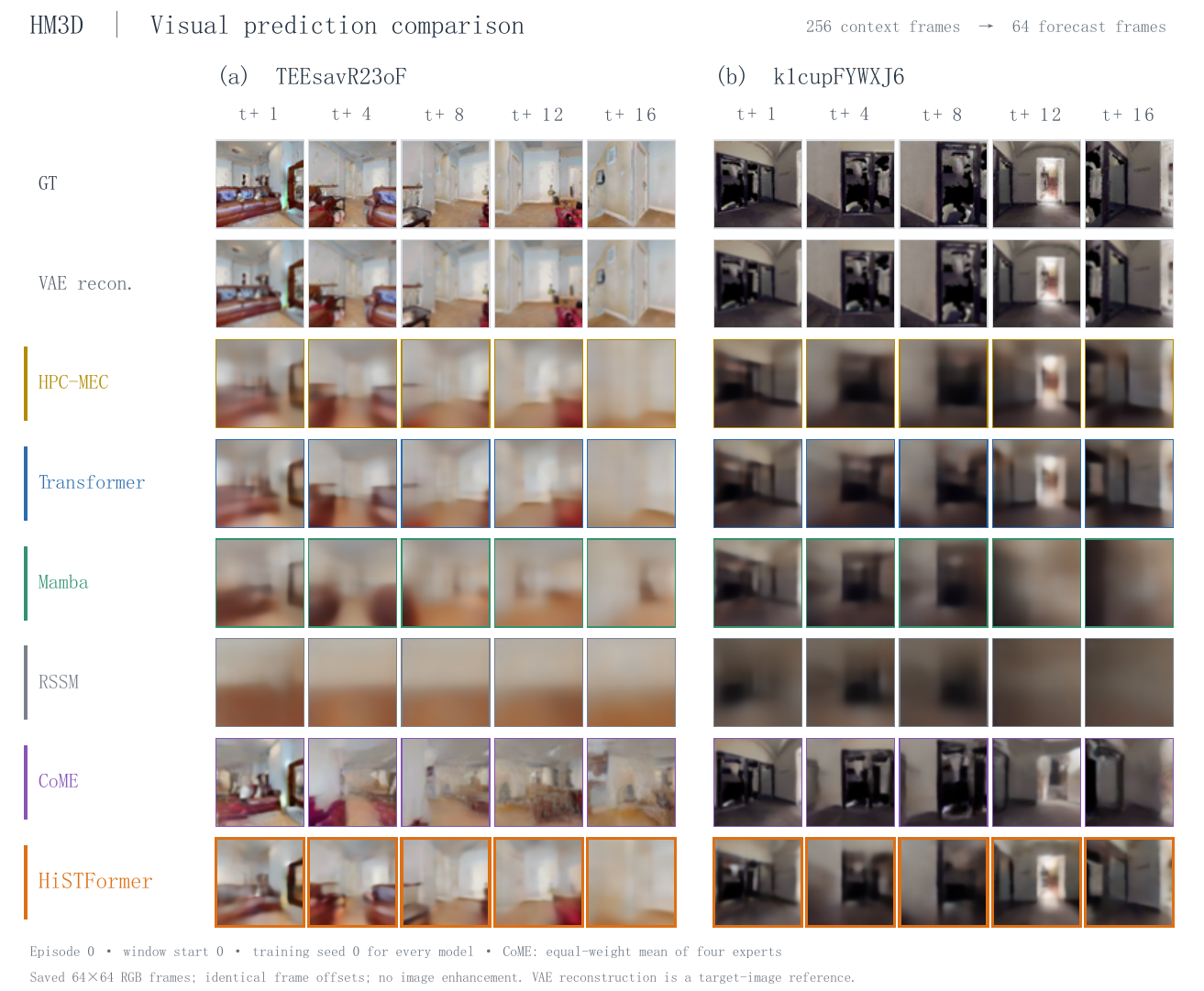}
    \caption{Qualitative predictions on HM3D-based trajectories. HiSTFormer
    preserves major scene structure and viewpoint evolution across multiple
    test scenes. The 32- and 64-frame predictions extend beyond the 16-frame
    training horizon; frozen-VAE reconstructions are shown as a reference for
    representation quality.}
    \label{fig:hm3d_adaptation}
\end{figure*}

Figure~\ref{fig:hm3d_adaptation} shows qualitative comparisons across different models and test scenes. HiSTFormer captures the major spatial layouts and viewpoint changes,
although fine details remain blurrier than the frozen VAE reconstructions.
When rollouts extend beyond the 16-frame training horizon to 32 and 64 frames,
recognizable structure persists in some scenes. In comparison, CoME consistently produces better predictions for the first 1–4 frames but exhibits more pronounced errors over longer rollouts. Transformer and HPC--MEC maintain relatively coherent predictions, although some frames lose more visual details than HiSTFormer, potentially indicating HiSTFormer's more effective use of historical observations. These results provide
preliminary evidence that HiSTFormer can adapt to complex indoor environments.

\section{Baseline Implementation Details}
\label{app:baseline_details}
The following implementation details are based on the MemoryMaze setting. For HM3D, we retain the same overall configurations, with adjustments only to model capacity and input data characteristics.

\subsection{HPC--MEC}
HPC--MEC uses the same visual encoding framework and motion-processing module
as HiSTFormer. The main differences lie in the MEC component and memory mechanism,
both of which follow the design of the original method.

For the MEC component, we compare two variants: \textit{(i)} continuously
propagating the MEC state within each chunk, and \textit{(ii)} re-inferring the
MEC state from the HPC state at every step. Based on a preliminary single-seed
experiment, we use the second variant in the reported experiments. During
open-loop prediction, the HPC state is driven by the model's predicted visual
representations and does not access future ground-truth images.

\subsection{CoME}
The main text reports CoME under a training budget approximately matched to
that of HiSTFormer. Specifically, ImageVAE, Base, STM, and Spatial Base are trained
for $20{,}000$, $10{,}000$, $30{,}000$, and $30{,}000$ gradient steps,
respectively. Following the original setup, ImageVAE is frozen after training.
LTM and SLTM perform online LoRA adaptation using the observed context of each
trajectory.

Extended-training results are reported in
Appendix~\ref{app:extended_baselines}. Under the
$500\!\rightarrow\!500$ evaluation protocol, evaluating a single sequence
takes approximately 278 seconds.

\subsection{General-Purpose World Models}
Transformer, RSSM, and Mamba take the previous visual latent, the current
transition action, and the ground-truth head direction at the target time as
inputs. Each model updates its latent state and autoregressively predicts the
subsequent visual latents.

The Transformer baseline is a two-layer causal model with eight attention heads
and a hidden width of 480. It uses learnable absolute positional embeddings
along the temporal axis. The Mamba baseline uses three Mamba-1 layers with a
hidden width of 464. The RSSM baseline uses a 768-dimensional deterministic
state together with a 234-dimensional continuous Gaussian latent state. All
three models contain approximately 8 million parameters.

Because learnable absolute positions may be unfavorable for extrapolation
beyond the training sequence length, we additionally consider a more
controlled relative-position comparison. The TT configuration in
Figure~\ref{fig:temporal_spatial_analysis}A is a two-layer temporal-only
Transformer using relative temporal RoPE and the same 128-token memory budget
as the other positional-encoding variants. Despite removing the absolute-position
extrapolation issue, TT still performs poorly under long-horizon evaluation,
particularly the $500\!\rightarrow\!500$ protocol. Thus, replacing absolute
embeddings with temporal RoPE alone does not yield strong long-horizon
generalization; the gains of HiSTFormer additionally rely on explicit spatial
indexing and layer-wise temporal--spatial organization.

\section{Extended Baseline Experiments}
\label{app:extended_baselines}

\subsection{HPC--MEC}
The original HPC--MEC formulation combines observation feedback with recurrent
MEC-state propagation. Our setting instead requires continuous open-loop
prediction after a fixed observation history, making it necessary to specify
how the MEC state is updated when no new image observations are available. We
therefore compare two update protocols: CI-R, which preserves recurrent
MEC-state propagation, and SI, which re-estimates the MEC state at every step
from the current HPC state. Under SI, the HPC state is updated autoregressively
using the model's own predicted visual representations and never accesses
future ground-truth images.

Both variants are trained from scratch for $10{,}000$ gradient steps using
identical training configurations and the same random seed. As shown in
Table~\ref{tab:hpc_mec_update_protocols}, SI achieves lower MSE under both the
short- and long-horizon protocols and consistently improves the long-horizon
perceptual metrics. We therefore adopt SI in the main experiments, while
reporting CI-R to make the effect of this implementation choice explicit.

\begin{table}[!ht]
\centering
\small
\setlength{\tabcolsep}{4pt}
\renewcommand{\arraystretch}{1.12}
\caption{Comparison of HPC--MEC update protocols. Both variants are trained
from scratch for $10{,}000$ gradient steps using identical configurations and
the same random seed. LPIPS, SSIM, and PSNR are reported under the
$500\!\rightarrow\!500$ protocol.}
\label{tab:hpc_mec_update_protocols}
\resizebox{\textwidth}{!}{%
\begin{tabular}{lrrrrr}
\toprule
MEC update & MSE ($256\!\rightarrow\!64$) $\downarrow$
& MSE ($500\!\rightarrow\!500$) $\downarrow$
& LPIPS $\downarrow$ & SSIM $\uparrow$ & PSNR $\uparrow$ \\
\midrule
CI-R & 0.012053 & 0.018827 & 0.410430 & 0.647472 & 17.2523 \\
SI & \textbf{0.009785} & \textbf{0.015406} & \textbf{0.369950}
& \textbf{0.662986} & \textbf{18.1231} \\
\bottomrule
\end{tabular}
}
\end{table}

\subsection{CoME}
The original CoME study uses a substantially larger training budget, with up
to one million gradient steps. To examine the effect of additional
optimization, we extend the training of Base, STM, and Spatial Base to
$30{,}000$, $50{,}000$, and $50{,}000$ gradient steps, respectively, while
keeping the remaining setup unchanged. We report results under the original
$200\!\rightarrow\!17$ protocol as well as the
$256\!\rightarrow\!64$ and $500\!\rightarrow\!500$ protocols used in our main
evaluation.

As shown in Table~\ref{tab:come_extended_training}, the extended budget improves
all four prediction metrics across the three protocols. Nevertheless, CoME
does not surpass HiSTFormer under either of the main evaluation protocols. CoME
also contains more than 293 million trainable parameters, substantially more
than HiSTFormer. Thus, even with additional optimization and a much larger model,
the performance gap is not closed, indicating lower training efficiency than
HiSTFormer.

\begin{table}[!ht]
\centering
\small
\setlength{\tabcolsep}{5pt}
\renewcommand{\arraystretch}{1.12}
\caption{CoME performance under the matched and extended training budgets.
The extended setting trains Base, STM, and Spatial Base for $30{,}000$,
$50{,}000$, and $50{,}000$ gradient steps, respectively.}
\label{tab:come_extended_training}
\begin{tabular}{llrrrr}
\toprule
Protocol & Training setting & MSE $\downarrow$ & PSNR $\uparrow$
& SSIM $\uparrow$ & LPIPS $\downarrow$ \\
\midrule
$200\!\rightarrow\!17$ & Matched budget
& 0.006815 & 21.672 & 0.6869 & 0.1143 \\
& Extended training
& \textbf{0.005252} & \textbf{22.797} & \textbf{0.7257} & \textbf{0.0927} \\
\midrule
$256\!\rightarrow\!64$ & Matched budget
& 0.017093 & 17.672 & 0.5738 & 0.2345 \\
& Extended training
& \textbf{0.015293} & \textbf{18.155} & \textbf{0.5954} & \textbf{0.2150} \\
\midrule
$500\!\rightarrow\!500$ & Matched budget
& 0.026328 & 15.797 & 0.4880 & 0.3169 \\
& Extended training
& \textbf{0.025996} & \textbf{15.851} & \textbf{0.4909} & \textbf{0.3120} \\
\bottomrule
\end{tabular}
\end{table}

\subsection{Transformer with Spatial Content Inputs}
\label{app:transformer_spatial_content}
To determine whether HiSTFormer benefits merely from access to a grid-like spatial
representation, rather than from using it through spatial RoPE, we train two
additional Transformer variants. HEX96 + Full injects the same 96-dimensional,
path-integrated grid-like representation used by HiSTFormer into the token content,
whereas XY + Full injects the path-integrated Cartesian coordinates directly.
Both variants use relative temporal RoPE and a full KV cache, following a
standard Transformer setting and providing access to the complete history.
All results are averaged over three training seeds.

Table~\ref{tab:transformer_spatial_content} shows that spatial content inputs
provide modest improvements in MSE and short-horizon LPIPS over the original
Transformer. However, neither variant improves long-horizon LPIPS, and both
remain substantially behind HiSTFormer under both protocols. Thus, simply adding
grid-like or Cartesian position information to token content is insufficient.
The grid-like representation is most effective when used as a rotary spatial
address that directly structures attention.

\begin{table}[!ht]
\centering
\small
\setlength{\tabcolsep}{4pt}
\renewcommand{\arraystretch}{1.12}
\caption{Comparison of spatial RoPE with spatial representations injected as
Transformer content. HEX96 + Full and XY + Full use relative temporal RoPE and
a full KV cache. Values are means over three training seeds.}
\label{tab:transformer_spatial_content}
\resizebox{\textwidth}{!}{%
\begin{tabular}{lrrrr}
\toprule
Model & MSE ($256\!\rightarrow\!64$) $\downarrow$
& LPIPS ($256\!\rightarrow\!64$) $\downarrow$
& MSE ($500\!\rightarrow\!500$) $\downarrow$
& LPIPS ($500\!\rightarrow\!500$) $\downarrow$ \\
\midrule
HiSTFormer (ours) & \textbf{0.005850} & \textbf{0.212404}
& \textbf{0.013286} & \textbf{0.292006} \\
Transformer & 0.010017 & 0.288581 & 0.028135 & 0.406308 \\
HEX96 + Full & 0.009399 & 0.280305 & 0.025517 & 0.411359 \\
XY + Full & 0.009054 & 0.276214 & 0.025456 & 0.418372 \\
\bottomrule
\end{tabular}
}
\end{table}

\subsection{Additional Comparison with ESWM}
ESWM is a biologically inspired spatial world model that infers environmental structure by storing and integrating state–action transition memories ~\citep{ESWM2026}. To compare different spatial memory mechanisms for long-horizon visual prediction, we implement an adapted version of ESWM following the original paper and pseudocode, and evaluate it on MemoryMaze and HM3D. The original method constructs a sparse memory bank using a minimum spanning tree based on environment topology. Since this requires explicit topological information, we retain all observed transitions as episodic memory. For a controlled comparison, we match HiSTFormer's visual encoder–decoder, 16-step autoregressive training protocol, and parameter count.

Table~\ref{tab:eswm_comparison} presents quantitative comparisons. On MemoryMaze, HiSTFormer outperforms ESWM across all metrics under the 256→64 protocol, reducing LPIPS by approximately 19.8\%. Under 500→500, their performance is comparable, with HiSTFormer achieving lower LPIPS. On HM3D, where spatial revisits are more frequent, both models perform similarly under 256→64, with HiSTFormer achieving slightly better mean values across all four metrics. Under 500→500, ESWM achieves better results across all four metrics.

Historical revisit frequency may partly explain these differences. Under HM3D's 500→500 protocol, approximately 96\% of predicted frames have matching poses in the observed history, and 92\% have matching state–action transitions. Since ESWM retains all observed transitions, it may benefit more directly from repeated experiences. In contrast, HiSTFormer remains highly competitive under such extensive historical coverage \textbf{while using fixed-capacity spatial memory}.

\begin{figure*}[t]
    \centering
    \includegraphics[width=\textwidth]{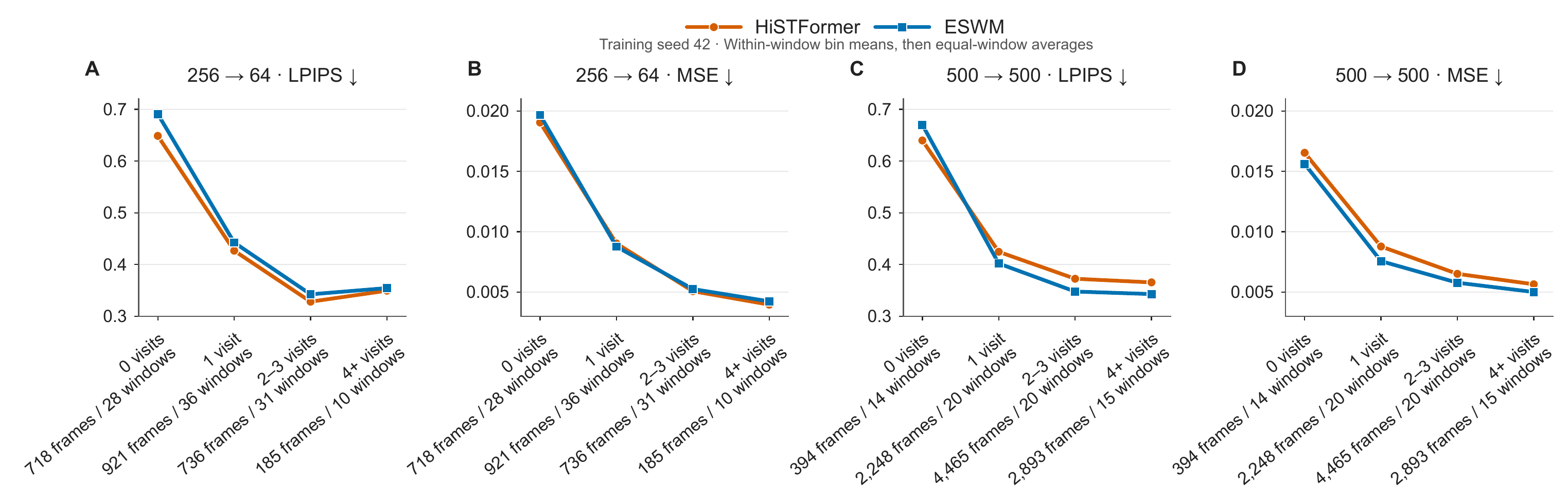}
    \caption{Visual Prediction Performance across Historical Revisit Frequencies. On HM3D, predicted frames are grouped by the number of historical pose-matching visits: 0, 1, 2--3, and 4 or more visits, using thresholds of 0.5 m in position and $30^\circ$ in heading. (A--B) LPIPS and MSE under the $256\!\rightarrow\!64$ protocol; (C--D) the corresponding metrics under $500\!\rightarrow\!500$. Both models are evaluated using the same training seed (Seed 42). Metrics are first averaged within each revisit group per test window, then averaged equally across windows containing that group. Prediction errors generally decrease with increasing revisit frequency for both models.}
    \label{app:hm3d_revisit_four}
\end{figure*}

Revisit-stratified analysis further reveals their different strengths. Figure ~\ref{app:hm3d_revisit_four} compares MSE and LPIPS across groups defined by the number of historical pose matches. Both models generally improve as revisit frequency increases. Under 256→64, HiSTFormer achieves lower LPIPS across all revisit groups and remains competitive in MSE. Under 500→500, ESWM performs better on samples with historical revisits. Notably, HiSTFormer achieves lower LPIPS under both protocols for frames without matching historical poses, indicating stronger perceptual prediction when direct historical correspondences are unavailable.

Overall, \textbf{HiSTFormer demonstrates a clear advantage under limited historical coverage while maintaining prediction quality comparable to a full-transition-memory model even in highly repetitive environments, despite its fixed memory capacity.} These findings further support the effectiveness of HiSTFormer's spatiotemporal organization and historical memory retrieval mechanisms.

\begin{table}[t]
\centering
\small
\setlength{\tabcolsep}{4pt}
\renewcommand{\arraystretch}{1.15}
\caption{Comparison of HiSTFormer and adapted ESWM on MemoryMaze and HM3D under two open-loop prediction protocols. Results are reported as mean $\pm$ standard deviation over three seeds. Best values are highlighted in bold.}
\label{tab:eswm_comparison}
\resizebox{\linewidth}{!}{%
\begin{tabular}{llcccc}
\toprule
Protocol & Model & MSE $\downarrow$ & LPIPS $\downarrow$ & SSIM $\uparrow$ & PSNR $\uparrow$ \\
\midrule
\multicolumn{6}{l}{\textbf{MemoryMaze}} \\
$256\!\rightarrow\!64$ & HiSTFormer & $\boldsymbol{0.005850 \pm 0.000250}$ & $\boldsymbol{0.2124 \pm 0.0027}$ & $\boldsymbol{0.7639 \pm 0.0027}$ & $\boldsymbol{24.723 \pm 0.160}$ \\
 & ESWM & $0.008594 \pm 0.000532$ & $0.2647 \pm 0.0068$ & $0.7287 \pm 0.0083$ & $22.997 \pm 0.356$ \\
$500\!\rightarrow\!500$ & HiSTFormer & $0.013286 \pm 0.000454$ & $\boldsymbol{0.2920 \pm 0.0037}$ & $0.7013 \pm 0.0047$ & $21.146 \pm 0.168$ \\
 & ESWM & $\boldsymbol{0.012702 \pm 0.000860}$ & $0.2973 \pm 0.0085$ & $\boldsymbol{0.7023 \pm 0.0093}$ & $\boldsymbol{21.288 \pm 0.370}$ \\
\midrule
\multicolumn{6}{l}{\textbf{HM3D}} \\
$256\!\rightarrow\!64$ & HiSTFormer & $\boldsymbol{0.01153 \pm 0.00035}$ & $\boldsymbol{0.4617 \pm 0.0018}$ & $\boldsymbol{0.5989 \pm 0.0017}$ & $\boldsymbol{21.448 \pm 0.064}$ \\
 & ESWM & $0.011556 \pm 0.000353$ & $0.484517 \pm 0.014441$ & $0.587461 \pm 0.009815$ & $21.316096 \pm 0.190845$ \\
$500\!\rightarrow\!500$ & HiSTFormer & $0.00784 \pm 0.00041$ & $0.4007 \pm 0.0093$ & $0.6422 \pm 0.0087$ & $22.367 \pm 0.203$ \\
 & ESWM & $\boldsymbol{0.006841 \pm 0.000278}$ & $\boldsymbol{0.378921 \pm 0.018619}$ & $\boldsymbol{0.663350 \pm 0.011288}$ & $\boldsymbol{22.865731 \pm 0.205280}$ \\
\bottomrule
\end{tabular}%
}
\end{table}


\section{Spatial Encoding Design and Robustness}
\label{app:spatial_encoding_design}
HiSTFormer adopts a biologically inspired multi-periodic grid-like encoding as its default spatial representation. This section examines two key design choices: multiple spatial periods and hexagonal plane-wave geometry. The ablations in the main text show that HiSTFormer is robust to different spatial encodings, with the grid-like representation achieving stable and generally favorable performance across the evaluated environment scales and prediction protocols. Here, we detail the encoding variants and further examine their theoretical motivations and empirical behavior.

\subsection{Spatial encoding variants.} 
For a controlled comparison, all three encodings use 96-dimensional representations and share the same path-integration procedure based on predicted displacements and ground-truth head directions. Only the encoding configurations differ:

\noindent\textbf{Grid-like (HiSTFormer):} Uses 8 logarithmically spaced periods, each containing 2 groups of 3 plane waves with directions separated by 120°. This yields 48 complex phases, corresponding to 96 real-valued dimensions.

\noindent\textbf{Cartesian: }Uses 2 orthogonal plane-wave directions (90°). To preserve the representation dimension, the number of periods is increased to 24, with 2 complex phases per period. The minimum and maximum periods match those of HiSTFormer, resulting in denser period sampling.

\noindent\textbf{Single-period}: Retains the three-direction grid-like geometry but fixes the spatial period to nearly 64 environment units. It uses 16 groups of 3 plane waves with randomized initial phase offsets, yielding 48 complex phases.

These variants examine the effects of wave geometry and period diversity under matched representation dimensions. Notably, Cartesian changes both the wave directions and period sampling density; therefore, this comparison does not isolate the effect of hexagonal geometry alone.

\subsection{Spatial Relevance of Memory Attention.}
To examine how spatial encoding affects historical retrieval, we compare the spatial attention distributions of Grid-like, Cartesian, and Single-period under the MemoryMaze $256\!\rightarrow\!64$ protocol. Only ground-truth observed memories are considered for spatial matching, excluding generated frames and the current token. Nearby locations are defined by a distance below 0.5 environment units, with a heading difference below $15^\circ$.

Table~\ref{tab:spatial_attention} shows modest advantages for Grid-like in spatially relevant retrieval. Nearby observations with matching headings receive 6.81\% of the attention within observed memories, compared with 6.51\% for Cartesian and 5.70\% for Single-period. Grid-like also achieves the shortest attention-weighted spatial distance (2.93 vs. 3.00/3.12). These results suggest slightly more spatially focused retrieval, consistent with downstream prediction performance, while confirming that HiSTFormer's effectiveness is not strongly dependent on a particular encoding geometry.

\begin{table}[t]
\centering
\small
\setlength{\tabcolsep}{5pt}
\renewcommand{\arraystretch}{1.15}
\caption{Spatial relevance of memory attention on MemoryMaze ($256\!\rightarrow\!64$). Nearby locations have a ground-truth distance below 0.5 environment units, and matching headings differ by less than $15^\circ$. All-attention fractions use the full attention distribution, while observed-memory fractions are normalized within ground-truth observed memories. Results are mean $\pm$ standard deviation over three seeds.}
\label{tab:spatial_attention}
\resizebox{\linewidth}{!}{%
\begin{tabular}{lccc}
\toprule
Metric & Grid-like & Cartesian & Single-period \\
\midrule
Observed memory / all attention (\%) & $53.13 \pm 2.16$ & $54.20 \pm 3.18$ & $55.75 \pm 1.80$ \\
Nearby + heading / all attention (\%) $\uparrow$ & $\mathbf{4.79 \pm 0.10}$ & $4.64 \pm 0.33$ & $4.37 \pm 0.34$ \\
Nearby + heading / observed attention (\%) $\uparrow$ & $\mathbf{6.81 \pm 0.18}$ & $6.51 \pm 0.11$ & $5.70 \pm 0.12$ \\
Nearby / observed attention (\%) $\uparrow$ & $\mathbf{11.59 \pm 0.31}$ & $11.09 \pm 0.10$ & $10.23 \pm 0.11$ \\
Attention-weighted distance (cells) $\downarrow$ & $\mathbf{2.93 \pm 0.07}$ & $3.00 \pm 0.09$ & $3.12 \pm 0.08$ \\
\bottomrule
\end{tabular}%
}
\end{table}

\subsection{Multi-periodic encoding and long-range prediction.}
Periodic spatial codes can assign similar phases to distant locations, resulting in positional ambiguity. Previous theoretical studies have shown that combining multiple spatial periods can extend the effective encoding range and reduce large-scale localization errors~\citep{mathis2012optimal,stemmler2015connecting}. To examine the implications for visual prediction, we analyze the prediction-error differences between Single-period and HiSTFormer under the 256--64 protocol, grouped by the spatial distance between each predicted position and the last observed position.

\begin{figure*}[t]
    \centering
    \includegraphics[width=\textwidth]{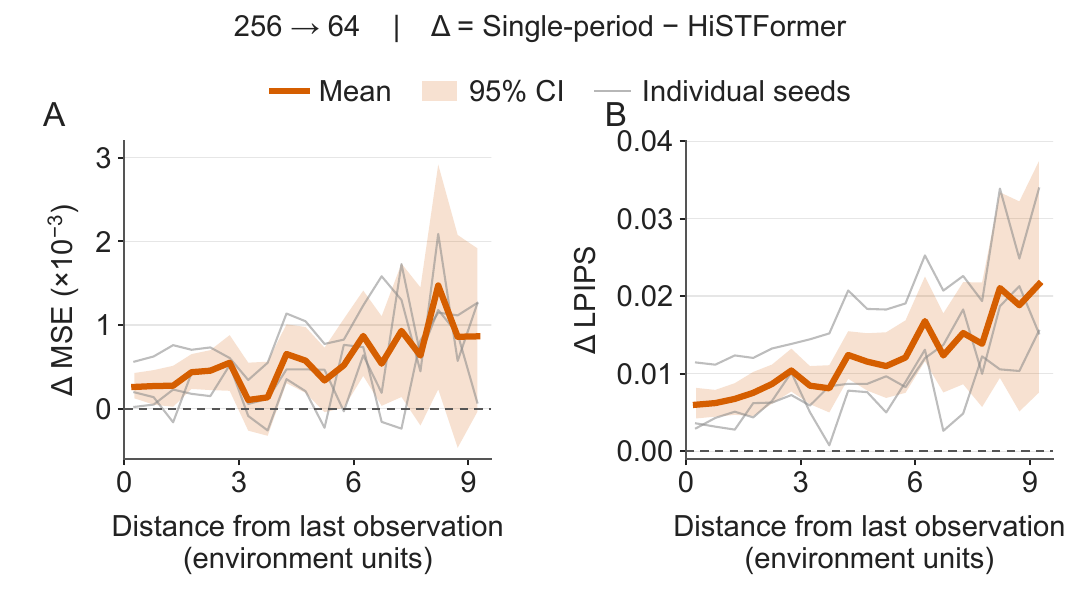}
    \caption{Distance-stratified prediction errors of single-period and multi-periodic spatial encodings. Under the 256-to-64 protocol, the horizontal axis shows the spatial distance from each predicted position to the last observed position. The vertical axis reports the error difference (Single-period minus HiSTFormer) in (A) MSE and (B) LPIPS. Orange curves show the mean over three training seeds, shaded regions indicate 95\% confidence intervals, and gray curves show individual seeds. Positive values indicate lower prediction errors for HiSTFormer.}
    \label{app:single-period}
\end{figure*}

Figure~\ref{app:single-period} reports the differences in MSE and LPIPS, computed as Single-period minus HiSTFormer, where positive values indicate lower prediction errors for HiSTFormer. The mean differences are generally positive and tend to increase with distance from the last observation. This trend is consistent with the theoretical expectation that single-period encodings are more susceptible to long-range phase ambiguity, providing further empirical support for multi-periodic spatial encoding.

\subsection{Motivation for hexagonal plane-wave geometry.}
Hexagonal spatial structure has a well-established biological basis, as observed in the firing patterns of entorhinal grid cells~\citep{hafting2005GridCell}. Theoretical and simulation studies also suggest potential geometric advantages. Guanella and Verschure reported favorable position reconstruction using triangular grid patterns compared with other regular two-dimensional tessellations~\citep{guanella2007prediction}. Schøyen et al. further showed that hexagonal grid patterns can offer improved spatial resolution and noise robustness over square patterns under conformal-isometry assumptions~\citep{schoyen2025hexagons}.

However, these advantages depend on specific encoding models and theoretical assumptions and do not necessarily translate into superior prediction performance in Transformer world models. Mir et al. found comparable performance between hexagonal and square grid representations in spatial memory storage and retrieval tasks~\citep{mir2024gridcells}. This agrees with our observations: although Grid-like achieves lower errors in some settings, its advantage over Cartesian is modest and less consistent, particularly under more challenging long-horizon prediction protocols.

\textbf{Overall, HiSTFormer's effectiveness is not tied to a particular spatial encoding.} Multi-periodic encoding is theoretically motivated by its ability to mitigate long-range positional ambiguity and shows corresponding empirical benefits in our experiments. Hexagonal plane-wave geometry provides a biologically grounded design with potential advantages under specific theoretical conditions. Considering these factors and the observed performance, we adopt the multi-periodic grid-like encoding as a reliable default, while emphasizing hierarchical spatiotemporal organization and content-adaptive spatial addressing as HiSTFormer's primary methodological contributions.
\begin{figure*}[t]
    \centering
    \includegraphics[width=0.7\textwidth]{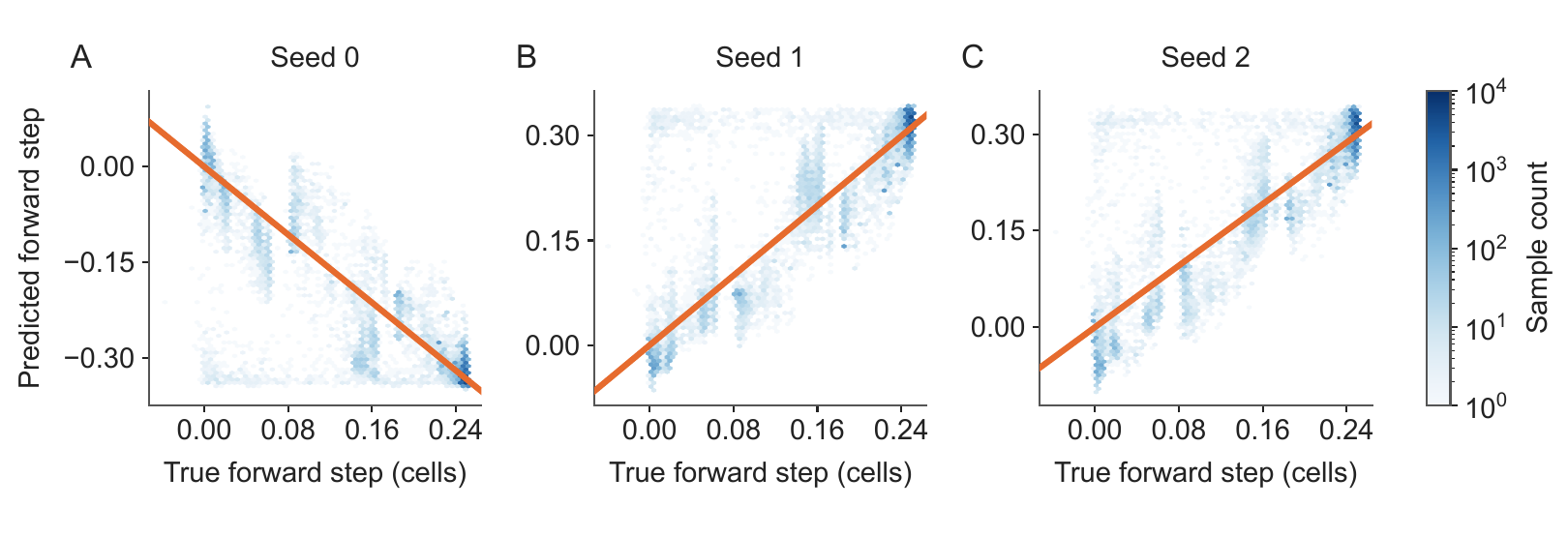}
    \caption{Motion Displacement Prediction Analysis. Comparison of forward displacements predicted by the action-only motion encoder and ground-truth forward steps across three training seeds. The horizontal and vertical axes represent ground-truth and predicted displacements, respectively, both in environment units. Blue hexagonal bins indicate sample density, with color intensity and opacity increasing logarithmically with sample count. Orange lines show linear fits. All three seeds exhibit approximately linear relationships, with differences in scale and sign.}
    \label{app:motion_pro}
\end{figure*}
\section{Interpretability Analysis of Content-Conditioned Spatial Addressing}
Content-conditioned offsets introduce only 0.051M additional parameters (approximately 0.67\%), reducing LPIPS by 9.00\% and 5.56\% under short- and long-horizon prediction, respectively. We further investigate their role through motion representations, head-specific offset patterns, and targeted memory interventions.

\subsection{The motion encoder learns spatially meaningful displacement representations}
We first examine the relationship between the forward displacement predicted by the action-only motion encoder and the ground-truth displacement. Figure~\ref{app:motion_pro} shows approximately linear relationships across three training seeds, with seed-dependent differences in scale and sign. This suggests that the encoder captures relative variations in physical motion. Since spatial phases are obtained by integrating predicted displacements, with content-conditioned offsets further adjusting these addresses, this result provides a basis for interpreting the learned spatial addressing geometrically.

\subsection{Attention heads learn stable, content-adaptive spatial offsets.} Figure~\ref{app:learn_offset}A shows the offset direction distributions of four heads across three seeds, together with the relationship between offset magnitude and the nearest-wall distance along each head's mean preferred direction. Most heads exhibit stable directional preferences, distributed approximately uniformly across the 80° field of view. Offset magnitudes are positively correlated with wall distances within approximately three environment cells, before gradually saturating. These results suggest that individual heads maintain directional preferences while adapting their retrieval distances to scene geometry. Figure~\ref{app:learn_offset}(B) presents three representative examples illustrating the alignment of learned offsets with visible walls.

\subsection{Spatial offsets exhibit geometric consistency with retrieved memories.} Figure~\ref{app:learn_offset}(C–E) further examines the representative case presented in the main text. In Figure~\ref{app:learn_offset}C, H2's adjusted address lies near the left red wall, while H0 points toward the distant green wall, albeit with some distance error. Figure~\ref{app:learn_offset}D shows historical frames receiving over 10\% attention from H0 or H2. H0 retrieves visual cues from the green wall, whereas H2 primarily attends to the red wall, suggesting complementary spatial information across heads.

\subsection{Targeted memory interventions further reveal the role of retrieved information.} 
\begin{figure*}[t]
    \centering
    \includegraphics[width=\textwidth]{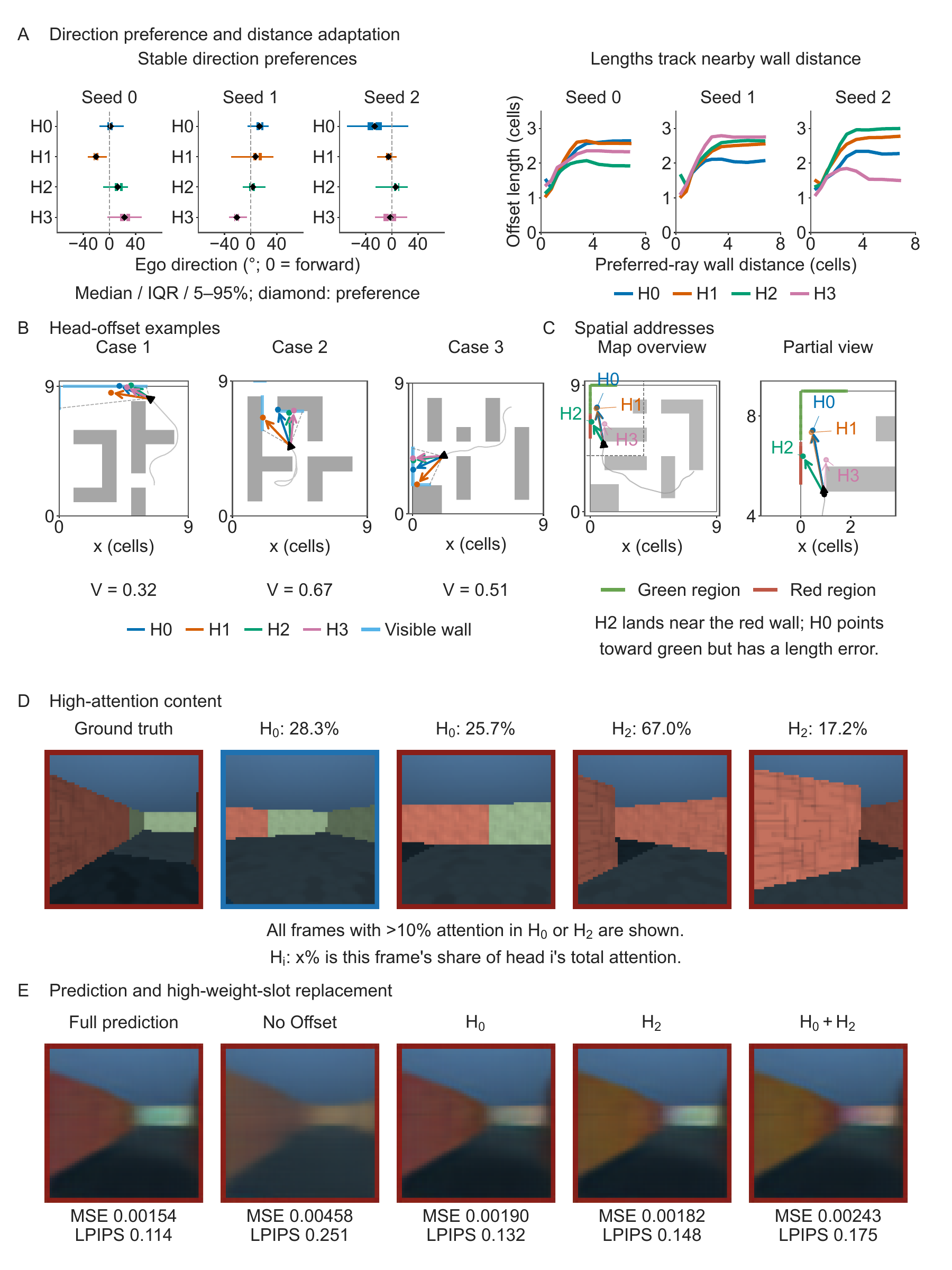}
    \caption{Geometric Alignment and Memory Retrieval of Content-Conditioned Spatial Offsets. (A) Head-specific directional preferences across three training seeds and the relationship between offset magnitude and nearest-wall distance along each head's preferred direction. Thin lines, thick lines, and white ticks indicate the 5th–95th percentile range, 25th–75th percentile range, and median, respectively. Black diamonds mark preferred directions estimated from separate calibration trajectories. (B) Head-specific spatial offsets in three representative scenes. (C) Spatial addresses and their relationships to the red and green walls in a representative case. (D) Historical memory frames receiving over 10\% attention from H0 or H2, with their attention weights. (E) Predictions from the full model, No Offset, and targeted replacement of high-attention memory slots in H0, H2, or both.}
    \label{app:learn_offset}
\end{figure*}

Figure~\ref{app:learn_offset}E compares the full model, No Offset, and predictions obtained by replacing high-attention memory slots in H0, H2, or both. The full model recovers the scene structure reasonably well, whereas No Offset exhibits clear errors in both walls. Replacing H0's high-attention slots primarily degrades reconstruction of the distant green region, while replacing H2's slots more strongly affects the left red wall. Intervening on both produces larger prediction errors. These changes align with the content attended by each head, providing further evidence that different heads exploit complementary spatial cues.

Overall, these analyses suggest that learned offsets establish geometrically meaningful, head-specific addressing patterns and adapt retrieval distances to scene content, enabling selective access to long-term memories and improving future visual prediction.

\end{document}